\documentclass[11pt, logo, nonumbering]{preprint}
\usepackage[utf8]{inputenc}
\usepackage{microtype}
\usepackage{graphicx}
\usepackage{amsmath}
\usepackage{import}
\usepackage{booktabs}
\usepackage{import}
\usepackage{multirow}
\usepackage[most]{tcolorbox}
 \usepackage{wrapfig}
\usepackage[T1]{fontenc}
\usepackage{xcolor}
\definecolor{mydarkblue}{rgb}{0,0.08,0.45}
\usepackage[colorlinks=true,linkcolor=mydarkblue,citecolor=mydarkblue,filecolor=mydarkblue,urlcolor=mydarkblue]{hyperref}
\usepackage{xspace}
\usepackage{cleveref}
\usepackage{CJK}

\usepackage[style=numeric-comp,maxbibnames=3,minnames=1,backend=bibtex, doi=false,eprint=false,url=false]{biblatex}

\definecolor{gred}{RGB}{250, 210, 207}
\definecolor{coolblue1}{rgb}{0.91, 0.94, 0.98}
\definecolor{coolblue2}{rgb}{0.76, 0.85, 0.94}
\definecolor{coolblue3}{rgb}{0.54, 0.72, 0.87}
\definecolor{coolblue4}{rgb}{1, 1, 1}

\newcommand{\method}{\textbf{WALAR}\xspace}

\begin{document}

\title{Mending the Holes: Mitigating Reward Hacking in Reinforcement Learning for Multilingual Translation}

\author{
\textbf{Yifeng Liu}$^{1}$ \quad
\textbf{Siqi Ouyang}$^{1}$ \quad
\textbf{Yatish H R}$^{1}$ \quad
\textbf{Lei Li}$^{1}$ \\
\textsuperscript{1}Carnegie Mellon University\\
\texttt{\{yifengl, siqiouya, yhosmane\}@andrew.cmu.edu} \\
        \texttt{leili@cs.cmu.edu}\\
    \href{https://github.com/LeiLiLab/WALAR}{\includegraphics[height=0.4cm]{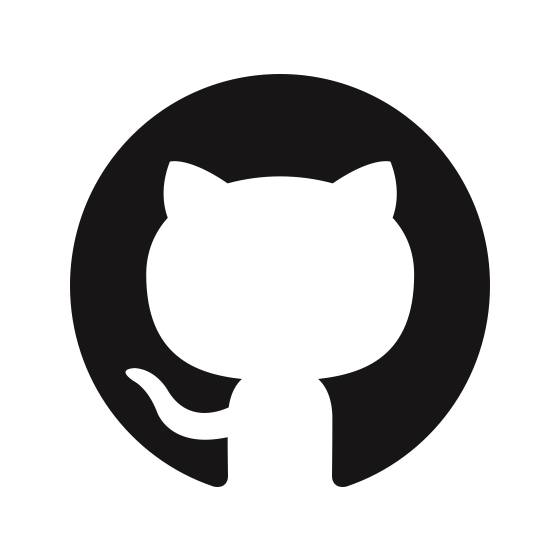} \textbf{LeiLiLab/WALAR}} ~ ~ ~ 
    \href{https://huggingface.co/collections/lyf07/walar}{\includegraphics[height=0.4cm]{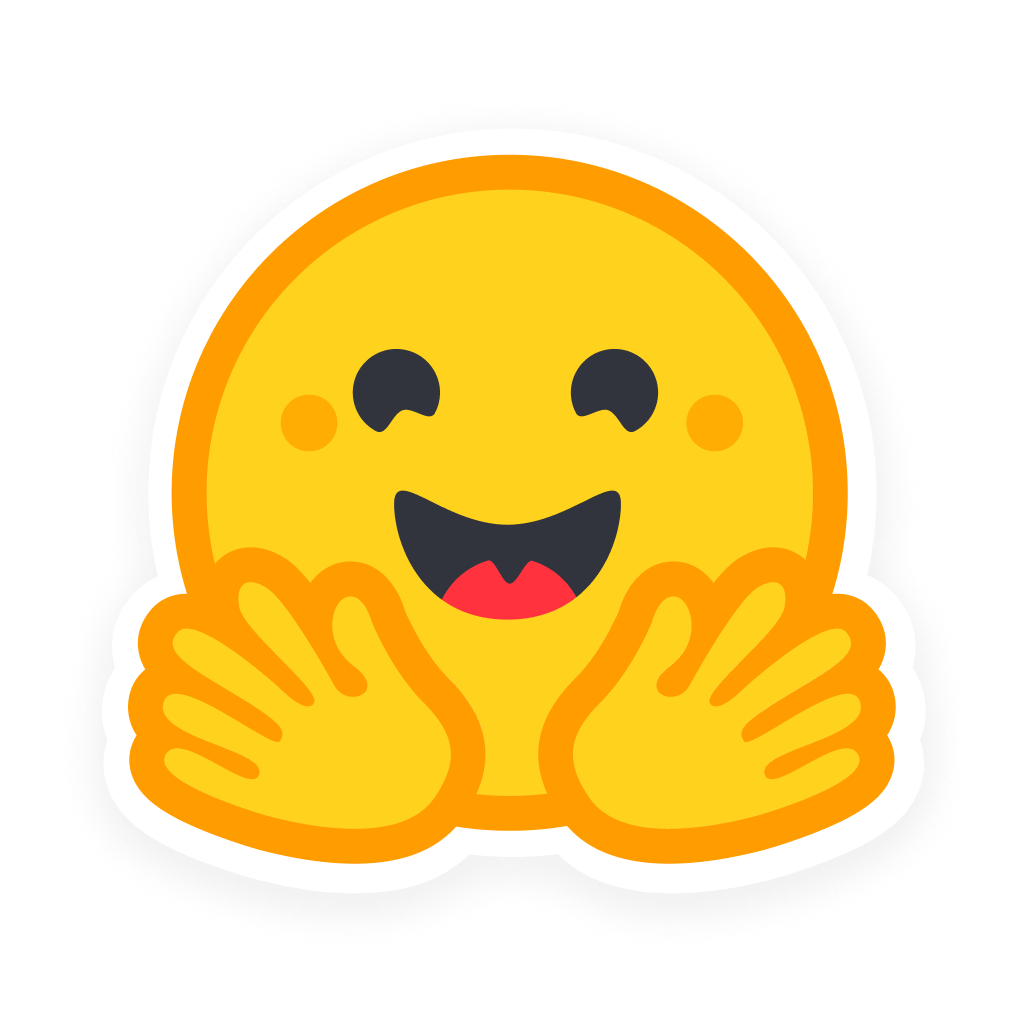} \textbf{lyf07/WALAR}}
}

\maketitle

\begin{abstract}
Large Language Models (LLMs) have demonstrated remarkable capability in machine translation on high-resource language pairs, yet their performance on low-resource translation still lags behind. 
Existing post-training methods rely heavily on high-quality parallel data, which are often scarce or unavailable for low-resource languages.
In this paper, we introduce \method, a reinforcement training method using only monolingual text to elevate LLMs' translation capabilities on massive low-resource languages while retaining their performance on high-resource languages. 
Our key insight is based on the observation of failure modes (or ``holes'') in existing source-based multilingual quality estimation (QE) models. 
Reinforcement learning (RL) using these QE models tends to amplify such holes, resulting in poorer multilingual LLMs. 
We develop techniques including word alignment and language alignment to mitigate such holes in \method's reward for RL training. 
We continually trained LLMs supporting translation of 101 languages using \method.  
The experiments show that our new model outperforms LLaMAX, one of the strongest open-source multilingual LLMs by a large margin on 1,414 language directions on \textsc{Flores-101} dataset\footnote{Our code is available at \url{https://github.com/LeiLiLab/WALAR}, and our models are available at \url{https://huggingface.co/collections/lyf07/walar}}. 
\end{abstract}

\section{Introduction}
\label{sec:intro}

Large Language Models (LLMs) exhibit strong capability on language translation, especially on high-resource language directions \cite{NEURIPS2020_1457c0d6,ouyang2022llmfollow,touvron2023llama,zhu-etal-2024-multilingual}. Recent progress in open source LLMs continuously pushes the quality of machine translation to a new level on par with human \cite{rei2025towerplus, grattafiori2024llama3herdmodels, yang2025qwen3technicalreport}. 
However, their translation performance on low-resource languages remains markedly inferior.\cite{zhu-etal-2024-multilingual,ochieng-etal-2025-beyond}.
Prior works on improving LLMs' translation capabilities focus primarily on post-training strategies such as supervised fine-tuning, knowledge distillation, and back-translation \cite{li2024elicitingtranslationabilitylarge, cheng2025seedxbuildingstrongmultilingual}. 
Despite the advancements, these methods are far from effective for low-resource or zero-resource languages since they rely on large amounts of high-quality parallel or preference data, which are scarce or unavailable for those languages.

\begin{figure}[!t]
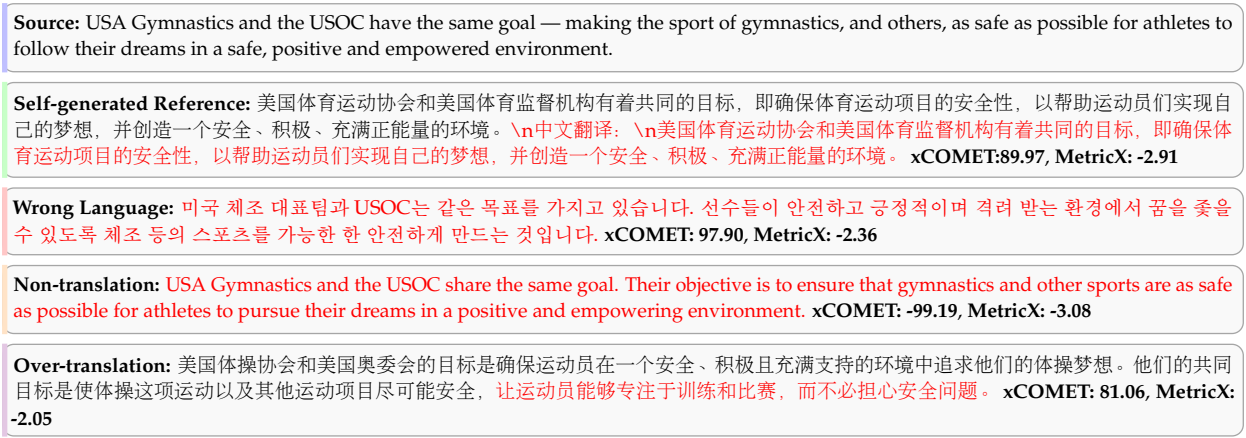

\begin{tcolorbox}[
    enhanced,
    width=\linewidth,
    colback=gray!5,
    colframe=gray!70,
    boxrule=0.5pt,
    arc=4pt,
    left=1pt,
    right=1pt,
    top=1pt,
    bottom=1pt,
    fontupper=\fontsize{7.5pt}{10pt}\selectfont,
    borderline west={2pt}{0pt}{blue!25}
]
\textbf{Source:} USA Gymnastics and the USOC have the same goal — making the sport of gymnastics, and others, as safe as possible for athletes to follow their dreams in a safe, positive and empowered environment.
\end{tcolorbox}
\vspace{-10pt}
\begin{tcolorbox}[
    enhanced,
    width=\linewidth,
    colback=gray!5,
    colframe=gray!70,
    boxrule=0.5pt,
    arc=4pt,
    left=1pt,
    right=1pt,
    top=1pt,
    bottom=1pt,
    fontupper=\fontsize{7.5pt}{10pt}\selectfont,
    borderline west={2pt}{0pt}{green!22}
]
\begin{CJK}{UTF8}{gbsn}
\textbf{Self-generated Reference:} 美国体育运动协会和美国体育监督机构有着共同的目标，即确保体育运动项目的安全性，以帮助运动员们实现自己的梦想，并创造一个安全、积极、充满正能量的环境。{\color{red}\textbackslash n中文翻译：\textbackslash n美国体育运动协会和美国体育监督机构有着共同的目标，即确保体育运动项目的安全性，以帮助运动员们实现自己的梦想，并创造一个安全、积极、充满正能量的环境。}
\textbf{xCOMET:89.97}, \textbf{MetricX: -2.91}
\end{CJK}
\end{tcolorbox}

\vspace{-10pt}
\begin{tcolorbox}[
    enhanced,
    width=\linewidth,
    colback=gray!5,
    colframe=gray!70,
    boxrule=0.5pt,
    arc=4pt,
    left=1pt,
    right=1pt,
    top=1pt,
    bottom=1pt,
    fontupper=\fontsize{7.5pt}{10pt}\selectfont,
    borderline west={2pt}{0pt}{red!22}
]
\begin{CJK}{UTF8}{mj}
\textbf{Wrong Language:} {\color{red}미국 체조 대표팀과 USOC는 같은 목표를 가지고 있습니다. 선수들이 안전하고 긍정적이며 격려 받는 환경에서 꿈을 좇을 수 있도록 체조 등의 스포츠를 가능한 한 안전하게 만드는 것입니다.}  
\textbf{xCOMET: 97.90}, \textbf{MetricX: -2.36}
\end{CJK}
\end{tcolorbox}
\vspace{-10pt}
\begin{tcolorbox}[
    enhanced,
    width=\linewidth,
    colback=gray!5,
    colframe=gray!70,
    boxrule=0.5pt,
    arc=4pt,
    left=1pt,
    right=1pt,
    top=1pt,
    bottom=1pt,
    fontupper=\fontsize{7.5pt}{10pt}\selectfont,
    borderline west={2pt}{0pt}{orange!22}
]
\begin{CJK}{UTF8}{mj}
\textbf{Non-translation:} {\color{red}USA Gymnastics and the USOC share the same goal. Their objective is to ensure that gymnastics and other sports are as safe as possible for athletes to pursue their dreams in a positive and empowering environment.} 
\textbf{xCOMET: -99.19}, \textbf{MetricX: -3.08}
\end{CJK}
\end{tcolorbox}
\vspace{-10pt}
\begin{tcolorbox}[
    enhanced,
    width=\linewidth,
    colback=gray!5,
    colframe=gray!70,
    boxrule=0.5pt,
    arc=4pt,
    left=1pt,
    right=1pt,
    top=1pt,
    bottom=1pt,
    fontupper=\fontsize{7.5pt}{10pt}\selectfont,
    borderline west={2pt}{0pt}{violet!22}
]
\begin{CJK}{UTF8}{gbsn}
\textbf{Over-translation:} 美国体操协会和美国奥委会的目标是确保运动员在一个安全、积极且充满支持的环境中追求他们的体操梦想。他们的共同目标是使体操这项运动以及其他运动项目尽可能安全，{\color{red}让运动员能够专注于训练和比赛，而不必担心安全问题。} 
\textbf{xCOMET: 81.06}, \textbf{MetricX: -2.05}
\end{CJK}
\end{tcolorbox}
\vspace{-10pt}
\caption{Holes of source-based quality estimation metrics. RL training using these metrics will amplify the holes in LLMs.}
\vspace{-8pt}
\label{fig:intro_case}
\end{figure}

We consider the following problem: can we effectively post-train an LLM with only monolingual data to improve translation performance on massive languages? 
Reinforcement learning (RL) has been applied effectively to improve standalone machine translation models and LLMs~\cite{kumar-etal-2019-reinforcement,yan-etal-2023-bleurt,he-etal-2024-improving,ramos-etal-2024-aligning}.  
The general idea is using a metric model such as COMET~\cite{rei-etal-2020-comet} or COMET-Kiwi~\cite{rei-etal-2022-cometkiwi} to provide reward signals during RL training. 
The former is reference-based --- comparing LLM's generation candidates to references --- while the latter is source-based. 
Since our scenario only contains monolingual text from multiple languages, we are forced to use source-based quality estimation (QE) models \cite{rei-etal-2022-cometkiwi,juraska2024metricx24googlesubmissionwmt}.

However, directly applying RL on LLMs with quality-estimation rewards presents notable weaknesses. Our study shows that, although state-of-the-art quality estimation models achieve strong performance in evaluating translation quality \cite{freitag-etal-2024-llms}, these QEs exhibit noticeable holes when applied to LLM training, such as failure to detect over- and under-translation, and wrong language words. 
Figure~\ref{fig:intro_case} illustrates examples of MetricX's inability to score major translation errors. 
Even worse, when trained with such QE rewards, an LLM could amplify holes in certain language directions, leading to reward hacking and resulting in the LLM just repeating input source sentences. 
Astonishingly, an QE model will give a perfect score to the generated repeating source when compared to the source utterance.

To solve this major challenge, we develop \method, an effective reinforcement learning method using monolingual-only data to enhance a pre-trained LLM's multilingual translation performance. 
Our key idea is to use a source-based quality estimation model as the base RL reward and to mitigate its holes with additional word alignment and language alignment scores. 
Word alignment will encourage proper coverage, not too many left or extra words in the candidate, compared to the source utterance. 
Language alignment will ensure the model is generating desired target languages. 
We integrate all these three components in the group relative policy optimization (GRPO) training framework and post-train LLMs based on Qwen3-8B~\cite{qwen3technicalreport}, LLaMAX3-8B-Alpaca~\cite{lu-etal-2024-llamax} and Translategemma-4B-it~\cite{finkelstein2026translategemmatechnicalreport}.
The outcome and our contributions are as follows: 
\begin{itemize}[leftmargin=1em,itemsep=1pt]
    \item We discover holes (failure modes) in widely-adopted QE models (xCOMET, MetricX) and observe that LLMs trained with these QEs lead to reward hacking in translating certain languages. 
    \item We develop \method, a reinforcement learning method for post-training multilingual LLM with a hybrid reward to mitigate reward hacking. 
    \item We trained three LLMs using our \method. Our experiments demonstrate that our models outperforms the strongest prior LLM of the same size in 1,414 language directions on the \textsc{Flores-101} dataset. Furthermore, \method generalizes across languages, improving the quality of multilingual translation even for unseen language directions during training.
\end{itemize}

\section{Related Work}
\label{sec:related}

\noindent\textbf{Reinforcement Learning in Machine Translation}
Performing RL on a machine translation task is not a novel idea.
\cite{feng-etal-2025-mt-r1} employs a reference-based model as the reward in the reinforcement learning to incorporate reasoning into LLMs' translating behavior.  \cite{ramos2025finegrainedrewardoptimizationmachine} leverages xCOMET as the reward model to generate token-level rewards, thus bringing a more fine-grained feedback and offering more benefit over sentence-level feedback.
However, these works rely heavily on reference translation data.
Other efforts have investigated the use of QE models in this context.
\cite{ramos-etal-2024-aligning} explores the potential of using the QE model as a data filter, reward model, and decoding reranker, demonstrating notable improvements in translation quality, whereas
\cite{he-etal-2024-improving} adopts QE-based feedback training and introduces heuristic rules to penalize the overoptimization problem of QE models.
Closely related to this line of work, \cite{pombal2025addingchocolatemintmitigating} systematically studies metric interference, showing that reusing the same or related automatic metrics for quality-guided decoding can severely distort instance-level metric scores and reduce their agreement with human judgments.

\noindent\textbf{Multilingual LLMs}
Recent progress in LLMs has continuously increased the supporting language numbers of LLMs \cite{yang2025qwen3technicalreport, grattafiori2024llama3herdmodels, xu2025xalmaplugplay} and achieved promising results on high-resource languages \cite{rei2025towerplus, cheng2025seedxbuildingstrongmultilingual}. But the performance gap between high- and low-resource languages remains significant \cite{yuan2024vocabularysharingfacilitatesmultilingualism, zhu-etal-2024-multilingual}. Efforts to address such a gap either focus on the pre-training phase \cite{lu-etal-2024-llamax} or the post-training phase \cite{rei2025towerplus, cheng2025seedxbuildingstrongmultilingual}. 
However, post-training methods, including instruction tuning and preference optimization, fail short in low-resource languages due to the scarcity of high-quality parallel data \cite{tran2020crosslingualretrievaliterativeselfsupervised, dang-etal-2024-rlhf}. \method offers promising potential to address this problem by utilizing the abundant monolingual data in low-resource languages, thereby incentivizing LLMs' translation capabilities solely with monolingual data.

\section{Proposed Method}
\label{sec:method}
\begin{figure*}[!ht]
    \centering
    \includegraphics[width=0.8\textwidth]{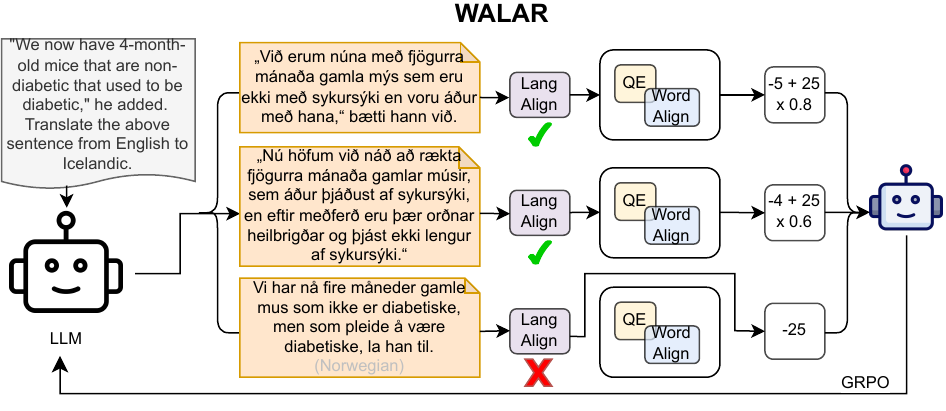}
    \vspace{-1em}
    \caption{Illustration of \method. On each step, the LLM is prompted to translate one monolingual sentence into another language with several different rollouts. Each output will then be evaluated by language alignment, quality estimation, and word alignment. Finally, the LLM is trained using GRPO with the reward on the previous step iteratively.}
    \label{fig:qe-rl}
\end{figure*}

In this section, we introduce the overall reinforcement training framework and our specially designed reward to mitigate hacking issues brought by translation quality estimation metrics.

\subsection{Problem Formulation}

Let a source-language sentence be represented as a sequence of tokens $x = (x_1, x_2, \ldots, x_m) \in L_{\text{src}}^m$, where $L_{\text{src}}$ denotes the source-language vocabulary and $m$ is the sequence length. A translation model (e.g., LLM) captures the conditional distribution of a target-language token sequence given the source sentence,
\begin{equation}
\pi_\theta(y \mid x) = \prod_{t=1}^{n} \pi_\theta(y_t \mid y_{<t}, x),
\end{equation}
where $y = (y_1, \ldots, y_n)$, $y_t \in L_{\text{tgt}}$, $L_{\text{tgt}}$ denotes the target-language vocabulary, $n$ is the target sequence length, and $\theta$ are the model parameters. 
We start from a pre-trained LLM and continually train it with only source text ($x$'s) in multiple languages using reinforcement learning (e.g., GRPO). 
It optimizes the following objective: 
\begin{align}
   \arg\!\max \mathcal{J}(\theta) = \mathbb{E}_{y \sim \pi_\theta(\cdot \mid x)}[R(x,y)]
\end{align}
where $y$ is sampled from prior model $\theta$ and $R$ is a carefully designed reward.
GRPO uses a slightly more sophisticated reward with an advantage function, which will be presented later. 


\subsection{\method Reward}

Our reward comprises three components: a base quality estimation model, word alignment score, and language alignment score. We first detail each component and then describe how they are integrated into a unified reward.

\paragraph{Quality Estimation Score.} To effectively evaluate the translation given only the source sentence, we use MetricX-24-Hybrid-XXL-Bf16\footnote{https://huggingface.co/google/metricx-24-hybrid-xxl-v2p6-bfloat16} (MetricX;  \cite{juraska2024metricx24googlesubmissionwmt}), the state-of-the-art quality estimation metric in WMT24 Metric Shared Task \cite{freitag-etal-2024-llms}. Remarkably, MetricX supports both source-based and reference-based evaluation as a hybrid model, achieving the highest consistency with human ratings. 
Besides, since MetricX is further finetuned from mT5 \cite{xue-etal-2021-mt5}, which is pretrained on mC4 and covers 101 languages, it can provide reliable evaluations even for translations into low-resource languages.

We define the QE reward $r_{\text{qe}}$ using MetricX as
\begin{equation}
r_{\text{qe}}(x, y) = \text{MetricX}(x, y),
\end{equation}
where the source sentence $x$ and LLM's generated hypothesis $y$ are concatenated with a separating space token and provided as input to the MetricX model to produce a scalar reward score $r_{\text{qe}}(x, y) \in [-25, 0]$, following the MQM annotation guidelines~\cite{juraska2024metricx24googlesubmissionwmt}.
However, using QE alone in RL would lead to reward hacking issues as we illustrated in Figure~\ref{fig:intro_case}, since QE may assign high rewards to degenerate hypotheses.

\noindent\textbf{Word Alignment Score.}
To address this reward hacking, we incorporate a word--alignment--based score that evaluates whether all words are properly covered in the target sentence and no extra information is introduced by LLM's hallucination.

Formally, a word aligner identifies a set of alignment pairs
\begin{equation}
\mathrm{WA} = \{(x_i, y_j) \mid x_i \in x,\; y_j \in y, \mathrm{Sim}(x_i, y_j)> c\},
\end{equation}
where each pair $(x_i, y_j) \in \mathrm{WA}$ indicates that the source token $x_i$ and the target token $y_j$ are semantically similar within the sentence context and $\mathrm{Sim}$ indicates semantic similarity.

We use the embedding-based approach from ~\cite{dou-neubig-2021-word} to calculate similarity and construct aligned word pairs in source-target utterances. 
Specifically, we first calculate the word embeddings $h_x=\langle h_{x_1},\ldots, h_{x_m}\rangle$ and $h_y=\langle h_{y_1},\ldots, h_{y_n}\rangle$ for $x$ and $y$ using an embedding model's hidden state. Then, we compute the similarity matrix through dot product $\mathrm{Sim}_\mathrm{xy}=Softmax(h_\mathrm{x}h_\mathrm{y}^T)$. 
We construct $\mathrm{WA}$ by taking the intersection: $\mathrm{WA}=\{(x_i, y_j)\mid \mathrm{Sim}_{xy}(x_i, y_j) > c \text{ and }  \mathrm{Sim}_{yx}(y_j, x_i) > c\}$, where $c$ is a threshold set to 1e-3. To ensure robustness in low-resource languages, we leverage BGE-M3, a strong multilingual embedding model supporting over 100 languages~\cite{chen-etal-2024-m3}, and extract word embeddings from its 24th layer.

Based on the constructed word alignments, we define the word-alignment score $r_{\text{wa}}$ as the F1 score:
\begin{equation}
r_{\text{wa}}(x, y) = 2 \cdot \frac{P(x, y) \cdot R(x, y)}{P(x, y) + R(x, y)},
\end{equation}
where $P(x, y) = \frac{|\mathrm{WA}|}{n}$ and $R(x, y) = \frac{|\mathrm{WA}|}{m}$ denote alignment precision and recall, respectively. This formulation penalizes both over-translation (which reduces precision) and under-translation (which reduces recall), thereby mitigating reward hacking effects induced by QE-based rewards.

\noindent\textbf{Language Alignment.}
Since both QE models and word alignment models are language-agnostic, LLMs can still hack theses scores by generating translations in an unintended language (see Section~\ref{sec:qe_holes}). 
To mitigate this issue, we introduce a language alignment score that verifies whether the generated translation matches the desired target language and only assigns a positive reward when the languages are as expected.

We adopt GlotLID~\cite{kargaran-etal-2023-glotlid}, a strong language identification model supporting over 1{,}600 languages, to detect the language of the LLM-generated translation. However, word alignment may assign disproportionately high scores when the translation copies words from the source sentence, which can lead to code-switching outputs after training. In our preliminary experiments, we find that GlotLID alone struggles to reliably identify such code-switching translations.

To address this limitation, we further incorporate MaskLID~\cite{kargaran-etal-2024-masklid}, a language identification method designed for code-switching scenarios. Specifically, we first apply MaskLID to detect code-switching segments in the generated translation. We then mask tokens belonging to these segments to obtain a filtered target sentence $y'$. Finally, we feed the masked sentence pair $(x, y')$ into GlotLID to compute the language-alignment reward $r_{\mathrm{la}}=\mathbb{I}(\mathrm{Lang\_{detect}}(y) =\mathrm{tgt})$, where $\mathrm{Lang\_detect}(\cdot)$ is the language detection function, $\mathrm{tgt}$ denotes the desired target language. This encourages the model to generate translations fully in the intended target language.

\noindent\textbf{Overall Reward.}
We define the overall \method reward function  as
\begin{equation}
r(x,y)=
\begin{cases}
-25, & \text{if  } r_{\mathrm{la}}=0\\

\begin{aligned}
\label{eq:overall_reward_function}
&r_{\mathrm{qe}}(x,y) +\alpha \cdot r_{\mathrm{wa}}(x,y'),
\end{aligned} & \text{if } r_{\mathrm{la}}=1
\end{cases}
\end{equation}
where $y'$ denotes the masked translation produced by the code-switching detector, and $\alpha$ is a scaling hyperparameter set to 20. We analyze the effect of $\alpha$ in Section~\ref{sec:analysis_alpha}.




\subsection{RL Training} 
We adopt Group Relative Policy Optimization (GRPO; \cite{shao2024deepseekmathpushinglimitsmathematical}) as our RL algorithm to train the model with our \method reward, as shown in Eq~\ref{eq:GRPO}.
\begin{equation}
\begin{aligned}
\label{eq:GRPO}
\mathcal{J}_{\mathrm{GRPO}}(\theta) 
&= \mathbb{E}_{x \sim D,\, \{y^{(k)}\}_{k=1}^G \sim \pi_{\theta_{\mathrm{old}}}(\cdot \mid x)} \\
&\Biggl[
  \frac{1}{G} \sum_{k=1}^G
  \min\!\Bigl(
    \frac{\pi_{\theta}(y^{(k)} \mid x)}{\pi_{\theta_{\mathrm{old}}}(y^{(k)} \mid x)}\, A_k,\, 
    \mathrm{clip}\!\bigl(
      \frac{\pi_{\theta}(y^{(k)} \mid x)}{\pi_{\theta_{\mathrm{old}}}(y^{(k)} \mid x)},
      1-\varepsilon,\,
      1+\varepsilon
    \bigr)
    A_k
  \Bigr) -\,\beta\,D_{\mathrm{KL}}\left(\pi_{\theta}\,\big\|\,\pi_{\mathrm{ref}}\right)
\Biggr],
\end{aligned}
\end{equation}


\begin{equation}
\label{eq:kl}
D_{\mathrm{KL}}(\pi_{\theta}\|\pi_{\mathrm{ref}}) = \frac{\pi_{\mathrm{ref}}(y^{(k)}|x)}{\pi_{\theta}(y^{(k)}|x)} - \log\frac{\pi_{\mathrm{ref}}(y^{(k)}|x)}{\pi_{\theta}(y^{(k)}|x)} - 1
\end{equation}

Specifically, for a query $x$ sampled from a monolingual dataset $D$, we first append a system prompt (``translating from language src to tgt'') to $x$. Then  
GRPO rolls out $G$ candidate sequences $\{y^{(1)}, y^{(2)}, \ldots, y^{(G)}\}$ at each step with old policy LLM $\pi_{\theta_{old}}$. 
For each  sequence, we extract the translation outputs (for simplicity, we slightly abuse x and y notations for modified input without and extracted translation from output).
For each output $y^{(k)}$, we compute the advantage $A_{k}=\frac{r(x, y^{(k)})-\mathrm{mean}(\{r(x, y^{(1)}), r(x, y^{(2)}), \ldots, r(x, y^{(G)})\})}{\mathrm{std}(r(x, y^{(1)}), r(x, y^{(2)}),\ldots, r(x, y^{(G)}))}$ with \method reward. 

The hyperparameters $\epsilon$ and $\beta$ control the GRPO clipping threshold and the weight of the Kullback--Leibler (KL) divergence penalty, respectively, in Eq~\ref{eq:kl}.

\section{Experiments}
\label{sec:experiment}
\subsection{Experimental Setup}
\label{sec:exp_setup}

\paragraph{Data.} Our monolingual training dataset is built upon the WMT News Crawl dataset \cite{kocmi-etal-2024-findings}, using 22 source languages\footnote{The source languages include:  Arabic, Bengali, Bulgarian, Croatian, German, English, Finnish, French, Hindi, Hungarian, Indonesian, Italian, Icelandic, Macedonian, Dutch, Polish, Portuguese, Romanian, Russian, Spanish, Turkish, Ukrainian, Simple Chinese.}. To effectively train the models, we first evaluate their performance with these 22 languages as the source and all other \textsc{Flores-101} languages supported by MetricX as the target. Then, we select language directions for which the sentence piece BLEU (spBLEU; \cite{goyal-etal-2022-flores}) score is between 1 and 20. Finally, for each selected language direction, we sample 250 instances and train all directions concurrently. In this way, we can avoid training models on language directions that are either too easy or too hard for them to translate, thus ensuring the effectiveness of our training process.
To ensure the quality of our training data, we adopt Named Entity Recognition (NER) and length clipping to filter out low-quality monolingual data. We also conduct data decontamination to avoid potential data leakage, following the approach in \cite{kocyigit2025overestimationllmevaluationcontrolled}.  For detailed information, please refer to Appendix~\ref{sec:data} and~\ref{app:lang_info}.

\paragraph{Models and training details.} Our implementation of \method is based on OpenRLHF\footnote{https://github.com/OpenRLHF/OpenRLHF} framework. During the training stage, we set the training batch size to 1024 and the micro-batch size to 16. For the GRPO algorithm, we set the rollout numbers to 8, the temperature to 1, the PPO clipping range $\epsilon$ to 0.2, and the KL penalty coefficient $\beta$ to 0.01. We also adopt warm-up training with the learning rate peaking at 5e-7. All the models are trained on 5 NVIDIA A6000 GPUs.
\label{sec:training_details}

We report results for strong multilingual encoder-decoder models and LLM-based decoder-only models. For the encoder-decoder model, we include NLLB-200-1.3B \cite{nllbteam2022languageleftbehindscaling}. For LLM-based decoder-only models, we evaluate Hunyuan-MT-7B \cite{zheng2025hunyuanmttechnicalreport}, Tower-Plus-9B \cite{rei2025towerplus}, Aya-Expanse-8B \cite{dang2024ayaexpansecombiningresearch}, Qwen3-8B in non-thinking mode \cite{qwen3technicalreport}, Translategemma-4B-it \cite{finkelstein2026translategemmatechnicalreport} and LLaMAX3-8B-Alpaca \cite{lu-etal-2024-llamax}, among which we further finetune LLaMAX3-8B-Alpaca, Qwen3-8B in non-thinking mode and Translategemma-4b-it with \textbf{\method}. Moreover, we employ another strong baseline LLaMAX3-8B-Alpaca+\textbf{\method-SFT}, which is a supervised fine-tuned model trained with high-scoring translations selected by \method's reward as pseudo-references. Specifically, we sample 32 possible translations for each sentence with min\_p=0.01 and select the translation with the highest WALAR's reward as the pseudo-reference. Then, we finetune LLaMAX-8B-Alpaca with the pseudo-references using cross entropy loss.

\begingroup
\renewcommand{\arraystretch}{1} 
\begin{table*}[!ht]
\centering
\footnotesize
\resizebox{\linewidth}{!}{
\begin{tabular}{l|cccccccccccccc|c}

\toprule
\textbf{spBLEU} & \multicolumn{1}{c}{\textbf{x $\rightarrow$ en}} & \multicolumn{1}{c}{\textbf{en $\rightarrow$ x}} & \multicolumn{1}{c}{\textbf{x $\rightarrow$ ar}} & \multicolumn{1}{c}{\textbf{ar $\rightarrow$ x}} & \multicolumn{1}{c}{\textbf{x $\rightarrow$ tr}} & \multicolumn{1}{c}{\textbf{tr $\rightarrow$ x}} & \multicolumn{1}{c}{\textbf{x $\rightarrow$ hi}} & \multicolumn{1}{c}{\textbf{hi $\rightarrow$ x}} & \multicolumn{1}{c}{\textbf{x $\rightarrow$ ru}} & \multicolumn{1}{c}{\textbf{ru $\rightarrow$ x}} & \multicolumn{1}{c}{\textbf{x $\rightarrow$ zh}} & \multicolumn{1}{c}{\textbf{zh $\rightarrow$ x}} & \multicolumn{1}{c}{\textbf{x $\rightarrow$ sw}} & \multicolumn{1}{c|}{\textbf{sw $\rightarrow$ x}} &\multicolumn{1}{c}{\textbf{Avg}}\\

\midrule
NLLB-200-1.3B$\Delta$ &39.03  & 30.23 & 24.91 & 22.57 & 23.30&22.47 & 24.51&22.22 &25.80 &22.18 & 18.71&18.40 &24.37 &21.77& 24.32  \\

Hunyuan-MT-7B  
& 21.04 & 14.32 
& 16.29 & 9.84 
& 15.37 & 10.72 
& 13.19 & 9.66 
& 16.17 & 10.46 
& 15.55 & 9.83 
& 8.18  & 7.46&12.72 \\

Tower-Plus-9B 
& 31.55 & 15.32 
& 9.42  & 10.69 
& 13.36 & 11.15 
& 20.53 & 11.39 
& 23.06 & 11.54 
& 23.01 & 9.98 
& 3.29  & 9.46&14.55 \\

Aya-Expanse-8B 
& 24.03 & 14.25 
& 14.59 & 10.33 
& 12.73 & 10.62 
& 14.07 & 10.20 
& 17.22 & 10.72 
& 15.49 & 9.29 
& 2.36  & 4.71&12.19 \\

\midrule
Qwen3-8B
& 29.22 & 16.67
& 15.70 & 11.11
& 14.00 & 11.45
& 12.10 & 11.48
& 19.05 & 12.08
& 21.08 & 10.58
& 1.29  & 6.72 & 13.75\\

\textbf{+\method} 
& 28.59 & 17.11
& 15.71 & 11.89
& 14.73 & 12.38
& 12.03 & 12.05
& 19.31 & 12.68
& 20.87 & 11.08
& 3.35  & 7.76 & 14.25\\

\midrule
Translategemma-4B-it 
& 27.45 & 18.19
& 18.86 & 12.58
& 16.39 & 13.05
& 17.85 & 12.75
& 20.62 & 13.74
& 19.46 & 11.39
& 10.17 & 11.61 &16.01 \\

\textbf{+\method}
& 28.26 & 19.81
& 19.06 & 13.70
& 16.65 & 14.34
& 17.76 & 14.41
& 21.11 & 14.75
& 19.90 & 12.21
& 12.44 & 13.24 &16.97 \\


\midrule
LLaMAX3-8B-Alpaca 
& 32.24 & 21.34 
& 17.50 & 14.14 
& 12.23 & 14.89 
& 16.64 & 15.42 
& 21.48 & 16.20 
& 18.37 & 12.99 
& 13.21 & 15.12&17.27 \\

\textbf{+\method-SFT}  
& \textbf{32.78} & 22.72 & \textbf{17.95} & 16.51 & 16.82 & 16.49 & \textbf{18.15} & 16.60 & 22.10 & 17.39 & 19.13 & 14.52 & 16.04 & 16.30&18.82 \\

\textbf{+\method}  
& 32.56 & \textbf{23.68} 
& 17.81 & \textbf{17.14}
& \textbf{18.02} & \textbf{17.86} 
& 18.00 & \textbf{17.61} 
& \textbf{22.45} & \textbf{17.90} 
& \textbf{20.08} & \textbf{15.23} 
& \textbf{17.15} & \textbf{17.35}&\textbf{19.49} \\
\bottomrule

\toprule
\textbf{xCOMET*} & \multicolumn{1}{c}{\textbf{x $\rightarrow$ en}} & \multicolumn{1}{c}{\textbf{en $\rightarrow$ x}} & \multicolumn{1}{c}{\textbf{x $\rightarrow$ ar}} & \multicolumn{1}{c}{\textbf{ar $\rightarrow$ x}} & \multicolumn{1}{c}{\textbf{x $\rightarrow$ tr}} & \multicolumn{1}{c}{\textbf{tr $\rightarrow$ x}} & \multicolumn{1}{c}{\textbf{x $\rightarrow$ hi}} & \multicolumn{1}{c}{\textbf{hi $\rightarrow$ x}} & \multicolumn{1}{c}{\textbf{x $\rightarrow$ ru}} & \multicolumn{1}{c}{\textbf{ru $\rightarrow$ x}} & \multicolumn{1}{c}{\textbf{x $\rightarrow$ zh}} & \multicolumn{1}{c}{\textbf{zh $\rightarrow$ x}} & \multicolumn{1}{c}{\textbf{x $\rightarrow$ sw}} & \multicolumn{1}{c|}{\textbf{sw $\rightarrow$ x}} & \multicolumn{1}{c}{\textbf{Avg}}\\

\midrule
NLLB-200-1.3B$\Delta$ &90.08 &79.00 &62.78 &70.50 &73.12 &72.77 &62.91 &69.44 &81.46 &76.23 &65.36 &70.49 &61.29 &61.93&71.24  \\
Hunyuan-MT-7B 
& 76.65 & 44.47 
& 52.51 & 39.38 
& 67.48 & 40.22 
& 55.90 & 38.29 
& 72.59 & 41.92 
& 67.69 & 41.72 
& 36.56 & 27.32&50.19 \\

Tower-Plus-9B 
& 82.03 & 45.13 
& 37.16 & 43.72 
& 55.09 & 42.99 
& 57.83 & 40.90 
& 76.78 & 45.00 
& 68.53 & 44.91 
& 20.36 & 35.95&49.74 \\

Aya-Expanse-8B 
& 70.89 & 45.04 
& 48.73 & 42.01 
& 53.54 & 42.02 
& 47.52 & 41.50 
& 66.00 & 44.36 
& 58.36 & 43.35 
& 18.83 & 22.45&46.04 \\
\midrule

Qwen3-8B
& 83.88 & 57.85
& 52.64 & 51.18
& 60.66 & 52.08
& 43.63 & 51.38
& 73.77 & 56.58
& 70.45 & 54.91
& 16.45 & 31.31 & 54.06\\

\textbf{+\method} 
& 84.77 & 62.98
& 55.15 & 56.07
& 64.55 & 57.13
& 48.27 & 55.78
& 76.71 & 61.17
& 71.37 & 59.61
& 21.84 & 35.17 &57.90 \\

\midrule
Translategemma-4B-it
& 87.10 & 65.40
& 56.63 & 57.41
& 71.01 & 58.77
& 60.93 & 56.22
& 81.25 & 62.73
& 72.57 & 60.97
& 44.62 & 48.64 &63.16 \\

\textbf{+\method}
& 87.73 & 70.04
& 57.39 & 61.72
& 72.25 & 63.11
& 60.25 & 62.38
& 81.48 & 67.26
& 72.60 & 65.38
& 51.43 & 53.00 &66.14 \\

\midrule
LLaMAX3-8B-Alpaca 
& 88.67 & 68.66 
& 57.61 & 61.38 
& 65.10 & 63.10 
& 53.50 & 61.34 
& 80.51 & 66.93 
& 70.29 & 64.80 
& 53.68 & 54.00&64.97 \\

\textbf{+\method-SFT}
& 89.16 & 71.48 
& 59.98 & 64.19 
& 68.43 & 66.13 
& 55.84 & 63.93 
& 81.11 & 69.99 
& 71.09 & 67.87 
& 56.09 & 56.45&67.27 \\

\textbf{+\method} 
& \textbf{90.44} & \textbf{76.42} 
& \textbf{68.06} & \textbf{68.48} 
& \textbf{73.36} & \textbf{71.05} 
& \textbf{58.56} & \textbf{68.03} 
& \textbf{82.39} & \textbf{74.65} 
& \textbf{71.99} & \textbf{72.80} 
& \textbf{62.20} & \textbf{60.31}& \textbf{71.34}\\

\bottomrule

\toprule
\textbf{MetricX*} & \multicolumn{1}{c}{\textbf{x $\rightarrow$ en}} & \multicolumn{1}{c}{\textbf{en $\rightarrow$ x}} & \multicolumn{1}{c}{\textbf{x $\rightarrow$ ar}} & \multicolumn{1}{c}{\textbf{ar $\rightarrow$ x}} & \multicolumn{1}{c}{\textbf{x $\rightarrow$ tr}} & \multicolumn{1}{c}{\textbf{tr $\rightarrow$ x}} & \multicolumn{1}{c}{\textbf{x $\rightarrow$ hi}} & \multicolumn{1}{c}{\textbf{hi $\rightarrow$ x}} & \multicolumn{1}{c}{\textbf{x $\rightarrow$ ru}} & \multicolumn{1}{c}{\textbf{ru $\rightarrow$ x}} & \multicolumn{1}{c}{\textbf{x $\rightarrow$ zh}} & \multicolumn{1}{c}{\textbf{zh $\rightarrow$ x}} & \multicolumn{1}{c}{\textbf{x $\rightarrow$ sw}} & \multicolumn{1}{c|}{\textbf{sw $\rightarrow$ x}}& \multicolumn{1}{c}{\textbf{Avg}} \\

\midrule

NLLB-200-1.3B$\Delta$ &-3.49 &-4.41 &-4.57 &-3.94 &-7.58 &-4.76 &-5.56 &-4.60 &-3.73 &-4.06 &-8.11 &-4.80 &-3.90 &-4.20 &-4.84  \\
Hunyuan-MT-7B  
& -7.88  & -14.91 
& -11.64 & -15.19 
& -10.13 & -15.36 
& -8.47  & -15.17 
& -8.38  & -15.13 
& -5.83  & -14.89 
& -14.90 & -17.38&-12.52 \\

Tower-Plus-9B 
& -6.13  & -15.00 
& -14.25 & -14.45 
& -12.80 & -14.87 
& -7.37  & -14.61 
& -7.25  & -14.73 
& -5.79  & -14.21 
& -19.83 & -15.90&-12.66 \\

Aya-Expanse-8B 
& -9.23  & -15.15 
& -13.98 & -15.24 
& -14.61 & -15.46 
& -10.64 & -14.36 
& -10.71 & -15.13 
& -9.20  & -15.08 
& -23.50 & -20.77&-14.50 \\

\midrule
Qwen3-8B
& -6.18  & -11.93
& -11.23 & -12.46
& -11.63 & -12.61
& -9.59  & -11.30
& -8.36  & -11.89
& -5.64  & -11.70
& -22.86 & -17.65 & -11.79\\

\textbf{+\method} 
& -5.84  & -10.13
& -10.53 & -10.38
& -10.51 & -10.64
& -9.02  & -9.49
& -7.48  & -10.06
& -5.27  & -9.76
& -21.01 & -15.81 &-10.42 \\

\midrule
Translategemma-4B-it
& -5.58 & -8.98
& -5.98 & -9.97
& -8.70 & -9.65
& -6.28 & -9.19
& -5.64 & -9.68
& -4.98 & -10.64
& -10.39 & -10.85 &-8.32 \\

\textbf{+\method}
& -5.16 & -6.76
& -5.77 & -7.05
& -7.91 & -7.39
& -6.01 & -6.37
& -5.20 & -7.17
& -4.44 & -8.51
& -8.49 & -8.22 &-6.75 \\



\midrule
LLaMAX3-8B-Alpaca 
& -4.11  & -7.74 
& -9.10  & -7.86 
& -9.65  & -8.07 
& -6.91  & -7.12 
& -5.52  & -7.50 
& -4.99  & -7.28 
& -9.39  & -8.66&-7.42 \\

\textbf{+\method-SFT}
& -4.01 & -6.92 & -8.46 & -6.89 & -8.56 & -7.08 & -6.31 & -6.28 & -5.33 & -6.59 & -4.77 & -6.24 & -8.56 & -7.75&-6.70\\

\textbf{+\method}  
& \textbf{-3.63}  & \textbf{-5.38} 
& \textbf{-6.03}  & \textbf{-5.46} 
& \textbf{-7.04}  & \textbf{-5.48} 
& \textbf{-5.68}  & \textbf{-4.99} 
& \textbf{-4.80} & \textbf{-5.17} 
& \textbf{-4.28}  & \textbf{-4.73} 
& \textbf{-6.78}  & \textbf{-6.12}&\textbf{-5.40} \\

\bottomrule

\toprule
\textbf{Gemini*}& \multicolumn{1}{c}{\textbf{x $\rightarrow$ en}} & \multicolumn{1}{c}{\textbf{en $\rightarrow$ x}} & \multicolumn{1}{c}{\textbf{x $\rightarrow$ ar}} & \multicolumn{1}{c}{\textbf{ar $\rightarrow$ x}} & \multicolumn{1}{c}{\textbf{x $\rightarrow$ tr}} & \multicolumn{1}{c}{\textbf{tr $\rightarrow$ x}} & \multicolumn{1}{c}{\textbf{x $\rightarrow$ hi}} & \multicolumn{1}{c}{\textbf{hi $\rightarrow$ x}} & \multicolumn{1}{c}{\textbf{x $\rightarrow$ ru}} & \multicolumn{1}{c}{\textbf{ru $\rightarrow$ x}} & \multicolumn{1}{c}{\textbf{x $\rightarrow$ zh}} & \multicolumn{1}{c}{\textbf{zh $\rightarrow$ x}} & \multicolumn{1}{c}{\textbf{x $\rightarrow$ sw}} & \multicolumn{1}{c|}{\textbf{sw $\rightarrow$ x}} & \multicolumn{1}{c}{\textbf{Avg}}\\

\midrule
LLaMAX3-8B-Alpaca &74.61 & 52.86 &51.14 &53.71 &57.18 &52.99 &61.04 &54.84 &66.08 &56.51 &64.18 &55.43 &50.80 &50.14& 57.25 \\
\textbf{+\method}  &\textbf{78.75} &\textbf{63.45} & \textbf{63.25}&\textbf{64.83} &\textbf{68.68} &\textbf{65.22} &\textbf{70.18} &\textbf{65.55} &\textbf{70.00} &\textbf{67.60} &\textbf{69.93} &\textbf{67.28} &\textbf{62.51} &\textbf{61.17}&\textbf{67.03}  \\

\bottomrule

\end{tabular}%
}

\caption{Model performance on \textsc{FLORES-101} test set, with results for 7 representative languages shown in the table. $\Delta$ denotes encoder-decoder models. Bold text denotes the best result across LLM-based decoder-only models. For spBLEU and Gemini*, we evaluate on all languages covered in \textsc{FLORES-101}. For xCOMET* and MetricX*, we only evaluate on the languages they support in \textsc{FLORES-101}.}
\label{tab:main_tab}
\end{table*}
\endgroup

\paragraph{Evaluation method.}
We evaluate all models on the \textsc{Flores-101} \cite{goyal-etal-2022-flores} test set using the BenchMAX evaluation suite \cite{huang-etal-2025-benchmax}, and report results for seven representative languages, covering 1,414 language directions in total. 
We use spBLEU \cite{goyal-etal-2022-flores}, XCOMET-XL\footnote{https://huggingface.co/Unbabel/XCOMET-XL} \cite{guerreiro-etal-2024-xcomet}, MetricX-24-Hybrid-XXL-Bf16 \cite{juraska2024metricx24googlesubmissionwmt} and Gemini 3 Flash \cite{geminiteam2025geminifamilyhighlycapable} to evaluate the translation quality of the models. 
To prevent LLMs from exploiting the neural metrics by generating wrong language translations (Section~\ref{sec:qe_holes}), we adopt GlotLID to identify the language of each translation candidate. Candidates identified as being in the wrong language are penalized by assigning the minimum score of the neural metric. We denote this penalized variant of xCOMET, MetricX and Gemini-based LLM-as-a-judge as xCOMET*, MetricX* and Gemini*, respectively. All three models are used in reference-based mode, with the source sentence, translation, and reference provided as inputs to ensure accuracy during evaluation. We evaluate xCOMET* and MetricX* only on languages they support, and spBLEU and Gemini* on all \textsc{Flores-101} languages. We also conduct human evaluation to further strengthen our results (Section~\ref{sec:human_eval}). Further details can be found in Appendix~\ref{sec:eval_detail}.

\subsection{Main Results}
\noindent\textbf{\method improves LLM translation quality by a large margin.}
As shown in Table~\ref{tab:main_tab}, we evaluate all models on the \textsc{Flores-101} benchmark and report spBLEU, xCOMET* and MetricX* scores over 1,414 language directions. Comparing Qwen3-8B, Translategemma-4B-it and LLaMAX3-8B-Alpaca before and after training with \method, we observe significant average improvements across all metrics, demonstrating the generalizability of \method across different model families.

Notably, \method yields substantial gains for both English-centric and low-resource-centric translation. For example, within the LLaMAX family, \method improves the xCOMET* score for Swahili-X from 54.00 to 60.31, and for English-X translation from 68.66 to 76.42. These significant improvements demonstrate the effectiveness of \method, particularly for low-resource language directions. We additionally provide the qualitative examples in Appendix ~\ref{app:qualitative_examples} and report the average rank across language pairs in Appendix~\ref{app:borda}.


\noindent\textbf{\method improves translation under LLM-as-a-Judge.}
To verify that \method improves actual translation quality rather than merely optimizing the neural metrics such as MetricX, we additionally evaluate translations using an LLM-as-a-Judge method. Specifically, we adopt Gemini 3 Flash as the judge model, motivated by the Gemini family's first-place performance in the WMT25 metrics shared task~\cite{lavie-etal-2025-findings}. Our evaluation prompt follows the ESA-style format used in WMT25, augmented with reference translations to enable reference-based assessment. The full prompt is provided in Appendix~\ref{appendix:llm-as-judge}.

As shown in Table~\ref{tab:main_tab}, we evaluate LLaMAX3-8B-Alpaca and its \method-trained counterpart on seven representative languages, covering over 1,400 language directions. Models trained with \method consistently outperform their baseline counterparts across all evaluated directions, increasing the average score from 57.25 to 67.03. Notably, the average score achieved by \method-trained LLaMAX3-8B-Alpaca is higher than 66, corresponding to translations with only minor issues according to the judging rubric. These results further corroborate the substantial translation quality improvements brought by \method.

\begin{figure}[!t]
    \centering
    \includegraphics[width=0.8\linewidth]{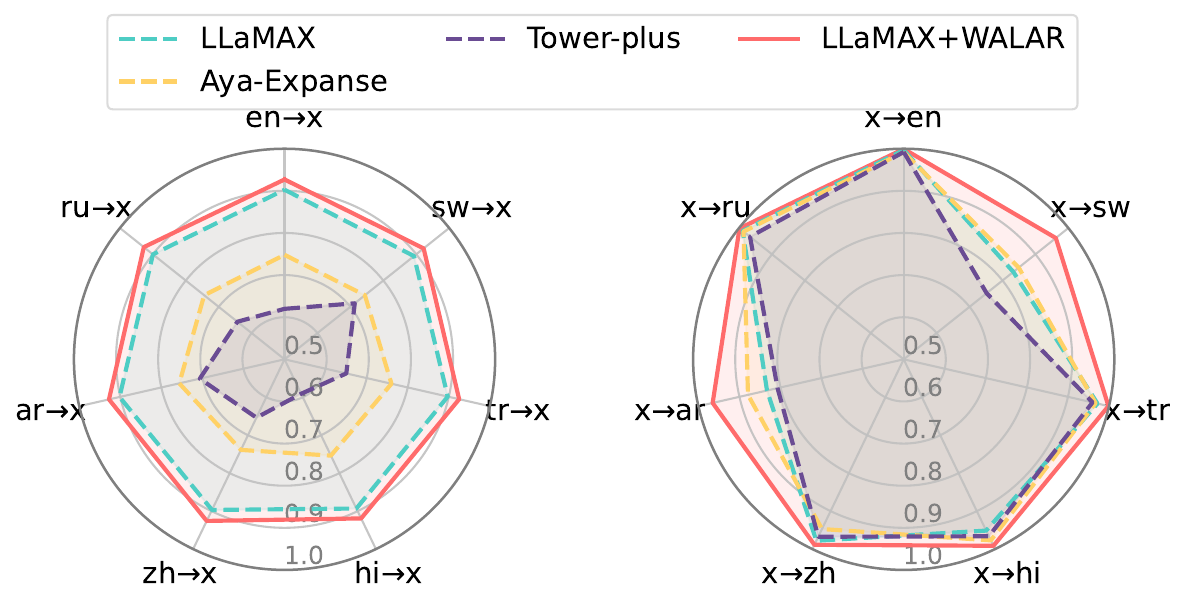}
    \caption{LCR on language directions.  \method improves LLMs' translation into desired target languages.}
    \label{fig:lcr}
\end{figure}

\paragraph{\method improves language consistency in translation.}
To systematically assess an LLM's ability to generate translations in the desired target language, we define the \emph{Language Consistency Rate} (LCR) as
\[
\mathrm{LCR} = \frac{\#\{\mathrm{Lang\_detect}(y) = \mathrm{tgt}\}}{\#\text{test data}},
\]
which measures the proportion of test instances whose outputs are identified as being in the correct target language. We report LCR for all language directions covered in Table~\ref{tab:main_tab}, using GlotLID~\cite{kargaran-etal-2023-glotlid} as the language identification model.

Figure~\ref{fig:lcr} presents the LCR results for four different decoder-only models. Training with \method consistently improves language consistency across all evaluated language directions on average. Among the four models, LLaMAX3-8B-Alpaca trained with \method achieves the highest LCR across all language directions. The improvement is particularly pronounced for low-resource target languages such as Swahili, where LCR increases from 83\% to nearly 100\%. Full results are reported in Table~\ref{tab:full_lcr}.




\section{Analysis}
\label{sec:analysis}

In this section, we present the analysis of \method and illustrate the holes of current neural machine translation metrics.

\subsection{Holes in Machine Translation Metrics}
\label{sec:qe_holes}
During training, we observe that models can exploit weaknesses in the reward signal when the reward itself is unreliable. Figure~\ref{fig:intro_case} summarizes the error types encountered during training. In particular, models trained solely with QE-based rewards exhibit several failure modes, including self-generated references, non-translation, over-translation, under-translation, and wrong language translation. Several of these failure modes are consistent with prior observations in the literature \cite{he-etal-2024-improving, yan-etal-2023-bleurt}.

\emph{Self-generated reference} refers to a failure mode in which the model learns to repeat its own hypothesis translation, causing the input to the QE model to take the form (source, hypothesis, hypothesis). This effectively tricks the QE model into treating the repeated hypothesis as a reference, activating its reference-based evaluation mode and yielding a high score. We attribute this behavior to the hybrid design of MetricX and xCOMET: during training, both models are optimized to support both source-based and reference-based evaluation by concatenating hypothesis translations and references into a single input.

\emph{Non-translation} occurs when the model simply paraphrases the source sentence rather than producing a translation. \emph{Wrong language translation} arises when the model generates output in a language different from the one specified in the prompt. In addition, models may exhibit \emph{over-translation} or \emph{under-translation}, producing outputs that contain redundant content or omit essential information.

We also provide the statistical analysis of each error category in Table~\ref{tab:hole_statistics}. Specifically, self-generated reference happens in our preliminary experiments on Qwen2.5-0.5B-Instruct with QE only reward. The model exhibits such behavior in 100\% cases. But it does not happen on larger base models like Qwen3-8B and LLaMAX3-8B-Alpaca. For wrong language translation, we measure the ratio of translation with wrong language for all four models with different reward configurations. Results are shown in the table below. The model trained with QE-only reward exhibits 92.43\% wrong language ratio as the QE model lacks the ability to tell whether the translation is in the right language direction. Language alignment score effectively fixes this issue. For over- and under-translation, we measure the average token length of the generated translations. \method is the only method whose translation length closely matches the reference, whereas other methods exhibit noticeable length deviation. This confirms that incorporating word alignment into the reward is critical to prevent both omission and over-generation.
\begin{table*}[]
\centering
\footnotesize
\resizebox{\linewidth}{!}{%
\begin{tabular}{@{}l|cccccc@{}}
\toprule
& \multicolumn{1}{c}{\textbf{LLaMAX}}& \multicolumn{1}{c}{+\textbf{\textit{QE}}} & \multicolumn{1}{c}{+\textbf{\textit{QE+Lang Align}}} & \multicolumn{1}{c}{+\textit{\method}} & \multicolumn{1}{c}{\textbf{Reference}} \\
\midrule
\textbf{Wrong Language Ratio (\%)} & 7.67 &92.43 & 3.96 &4.44 & 0.00  \\
\midrule
\textbf{Translation Token Length} &83.02  & 40.00 &79.09 & 65.24 & 62.04  \\
\bottomrule
\end{tabular}%
}
\caption{Statistics for different types of errors.}
\label{tab:hole_statistics}
\end{table*}

\begin{table}[t]
\centering
\footnotesize
\setlength{\tabcolsep}{3pt}
\begin{tabular}{lcccc}
\toprule
\multirow{2}{*}{\textbf{Model}} 
& \multicolumn{2}{c}{\textbf{Lang$\rightarrow$x}} 
& \multicolumn{2}{c}{\textbf{x$\rightarrow$Lang}} \\
\cmidrule(lr){2-3} \cmidrule(lr){4-5}
 & spBLEU & xCOMET* & spBLEU & xCOMET* \\
\midrule
LLaMAX3-8B-Alpaca & 15.73 & 65.73 & 18.81 & 69.46 \\
\midrule
+\textit{\textbf{QE}} (w/ filter) & 1.47 & 0.01 & 3.27 & 0.13 \\
+\textit{\textbf{QE+Lang Align}} (w/ filter) & 12.38 & \textbf{70.84} & 13.71 & \textbf{72.95} \\
+\textit{\textbf{\method}} (w/ filter) & \textbf{18.11} & 70.25 & \textbf{20.87} & 72.43 \\
+\textit{\method} (w/o filter) & 17.75 & 70.09 & 20.73 & 71.62 \\
\bottomrule
\end{tabular}
\caption{Ablation on the reward components of \method and spBLEU-based data filtering. \textbf{Lang} denotes the set of seven representative languages (English, Arabic, Turkish, Hindi, Russian, Swahili).}
\label{tab:ablation_study}
\end{table}
\subsection{Ablation Study}
We conduct the ablation study to demonstrate the contribution of each component in \method.
As shown in Table~\ref{tab:ablation_study}, we train the LLaMAX3-8B-Alpaca with three different rewards: (1) Quality estimation score, (2) Quality estimation score and language alignment, and (3) Quality estimation score, word alignment score and language alignment (\method). We also use \method to train the LLaMAX3-8B-Alpaca with data not filtered by spBLEU heuristics (Section~\ref{sec:exp_setup}). All models are trained with the same settings described in Section~\ref{sec:exp_setup} and evaluated on the same 1,414 language directions in Table~\ref{tab:main_tab}.

Results show that the LLaMAX trained with only the quality estimation score performs worst on both spBLEU and xCOMET*, primarily due to wrong language translations. Adding language alignment improves xCOMET* scores but degrades spBLEU, as it tends to over-translate. In contrast, \method achieves the best performance on both metrics, demonstrating the importance of word alignment score and language alignment. Additionally, LLaMAX3-8B-Alpaca, trained on all language directions, performs slightly worse than its counterpart trained with spBLEU-filtered language directions, which demonstrates the superiority of spBLEU-based data filtering.

\subsection{Effects of Hyperparameters}
\label{sec:analysis_alpha}
The hyperparameter $\alpha$ controls the weight of word alignment reward in \method (Eq~\ref{eq:overall_reward_function}). In this subsection, we focus on the question: How to select the best $\alpha$ for our model's training? To answer this question, we train the LLaMAX3-8B-Alpaca with six different $\alpha$: 0, 5, 10, 15, 20, 25, and evaluate all the checkpoints on \textsc{Flores-101} validation set with spBLEU, MetricX*, and xCOMET* as the metrics. As illustrated in Table~\ref{tab:alpha_tab}, by increasing $\alpha$ from 0 to 20, the spBLEU improves steadily from 12.88 to 19.71, while the MetricX* and xCOMET* degrade. $\alpha=25$ shows the worst performance across all three metrics. Finally, we report results for $\alpha=20$ as the hyperparameter in our experiments. We prioritize spBLEU for model selection for two reasons. First, spBLEU is more reliable for low-resource languages because it is a rule-based metric that relies on a fixed multilingual tokenizer rather than a learned neural model. Second, xCOMET and MetricX may be susceptible to over-optimization, since our training procedure directly optimizes toward neural metrics, which can lead to metric inflation. 

\begin{table*}
  \centering
  \resizebox{\textwidth}{!}{
\begin{tabular}{l|rrr|rrr|rrr}
\toprule
& \multicolumn{3}{c|}{\textbf{x $\rightarrow$ Lang (Avg.)}} & \multicolumn{3}{c|}{\textbf{Lang $\rightarrow$ x (Avg.)}} & \multicolumn{3}{c}{\textbf{Overall Avg.}}\\
\textbf{} & \textbf{spBLEU} & \textbf{MetricX*} & \textbf{xCOMET*} & \textbf{spBLEU} &\textbf{MetricX*} & \textbf{xCOMET*} & \textbf{spBLEU} &\textbf{MetricX*} & \textbf{xCOMET*} \\
\midrule
$\alpha$=0 & 13.44 & \textbf{-4.44} &\textbf{74.99}  & 12.31 & \textbf{-4.31} &72.17  &12.88 & \textbf{-4.38} & 73.58\\
$\alpha$=5 & 19.56 & -4.94 & \textbf{74.99} & 17.08 &-4.69 & \textbf{72.98} &18.32 &-4.82 &\textbf{73.99}\\
$\alpha$=10 & 20.54 & -5.45 & 74.42 & 17.88 &-4.88 & 72.46&19.21 &-5.17 &73.44\\
$\alpha$=15 & 20.98 & -5.32 & 73.92 & 18.12 & -4.97 & 71.78&19.55 &-5.15 &72.85\\
$\alpha$=20 & \textbf{21.10} & -5.44 & 73.39 & \textbf{18.32} & -5.10 & 71.66& \textbf{19.71}&-5.27 &72.53\\
$\alpha$=25 & 18.42 & -5.53 & 71.93 & 14.89 & -5.16 & 70.25&16.66 &-5.35 &71.09\\
\bottomrule
\end{tabular}
}
\caption{Performance of LLaMAX3-8B-Alpaca trained with different $\alpha$ on Flores-101 validation set. We select and report the results of $\alpha=20$ in all experiments.}
\label{tab:alpha_tab}
\end{table*}

\subsection{Human Evaluation}
\label{sec:human_eval}
As discussed in Section~\ref{sec:qe_holes}, neural metrics can be exploited by imperfect translations. To provide a more comprehensive evaluation beyond Gemini-based LLM-as-a-Judge on previous results, we conduct human evaluations on Azerbaijani-Portuguese (Az-Pt) and English-Kannada (En-Kn) translation tasks.

For each test instance, human annotators are presented with two translations, one generated by LLaMAX3-8B-Alpaca and the other by our \method-trained model, in a randomly permuted order. Annotators are asked to choose one of three options: (1) Translation~1 is better, (2) Translation~2 is better, or (3) Translation~1 and Translation~2 are of equal quality. We aggregate the annotations to compute win, loss, and tie rates. Additional details regarding the evaluation protocol are provided in Appendix~\ref{app:human_eval}.

Figure~\ref{fig:human_evaluation} summarizes the human evaluation results. Our model is preferred in 42\% of the cases for Az-Pt and 51\% for En-Kn, while producing translations of comparable quality in 34\% and 39\% of the cases, respectively. These results further corroborate the effectiveness of \method in improving translation quality, particularly for low-resource language pairs.



\begin{figure}[htbp]
  \centering
  \begin{minipage}[htbp]{0.5\linewidth}
    \centering
    \includegraphics[width=0.9\linewidth]{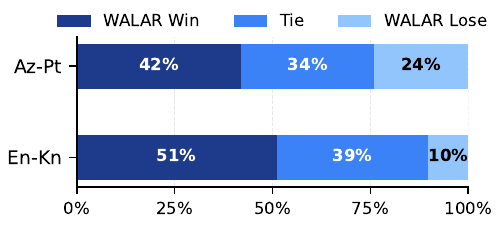}
    \captionof{figure}{Human evaluation results on Az-Pt and En-Kn.}
    \label{fig:human_evaluation}
      
  \end{minipage}\hfill
  \begin{minipage}[htbp]{0.5\linewidth}
  \centering
      \includegraphics[width=0.9\linewidth]{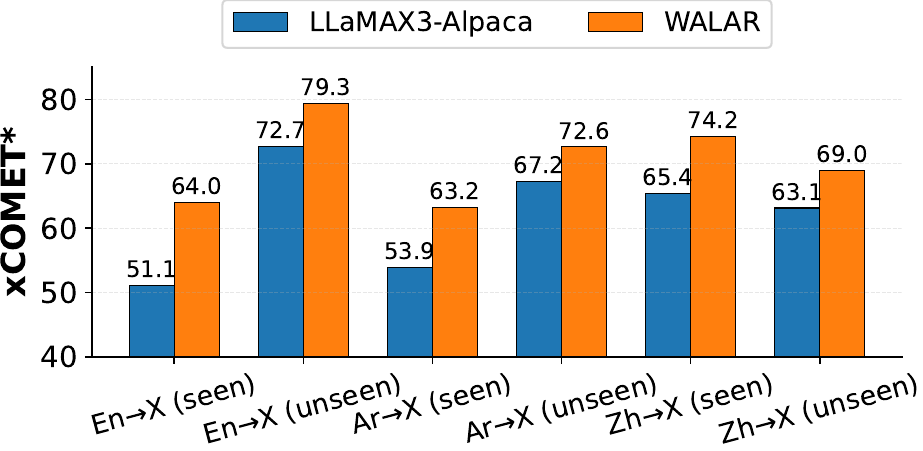}
    \captionof{figure}{Cross-lingual generalization on unseen target languages. X denotes languages in \textsc{Flores-101}. LLaMAX3-Alpaca, trained with \method, demonstrates strong generalization across unseen languages.}

  \label{fig:generalization}
  \end{minipage}
\end{figure}



\subsection{Generalization of \method}
\label{sec:generalization}
Despite the substantial improvements observed on \textsc{Flores-101} (Table~\ref{tab:main_tab}), an important question remains: can \method improve translation quality for unseen language directions when only monolingual data are available during training? To address this question, we evaluate LLaMAX3 and its \method-trained counterpart on 303 language directions (\{En, Ar, Zh\}$\rightarrow$x), and report results separately for seen and unseen target languages.

As shown in Figure~\ref{fig:generalization}, \method yields consistent gains on language directions observed during training, while also demonstrating strong cross-lingual generalization to unseen target languages. These results indicate that the improvements induced by \method can transfer beyond the training language set, potentially reducing the amount of parallel data and the number of language directions required to train large-scale multilingual models.

\section{Conclusion}
\label{sec:conclusion}
In conclusion, we present \method, a reinforcement training method that integrates quality estimation, word alignment, and language alignment as a reward to enhance LLM's translation ability in low-resource languages. Extensive experiments on \textsc{Flores-101} across 100 languages and over 1400 language directions show that \method enables LLMs to achieve substantial improvements on translation quality and language consistency. Our results on LLM-as-a-Judge and human evaluation further corroborate the effectiveness of \method. Finally, our analysis demonstrates the underexplored holes in current machine translation metrics and the generalization of \method to unseen languages during training.

\clearpage
\printbibliography

\newpage
\appendix
\label{sec:appendix}

\section{Data Curation}
\label{sec:data}
We collect all our monolingual data from the WMT News Crawl dataset \cite{kocmi-etal-2024-findings}, then perform data decontamination and data filtering for the source languages. Our data filtering process consists of two steps: length-based filtering and NER-based filtering.
\paragraph{Data Decontamination} We follow the method in ~\cite{kocyigit2025overestimationllmevaluationcontrolled} and implement an 8-gram search to find matches between our monolingual training dataset and \textsc{Flores-101} devtest data in corresponding languages. We tokenize the sentences into sub-word tokens and label the data as contaminated if the longest matching sub-sequence matches more than 70\% of the target tokens in \textsc{Flores-101} devtest.

\paragraph{Length-based Filtering} We directly use the tokenizer of Qwen3-8B to process \textsc{Flores-101}. Then, based on the token length distributions in each language, we empirically determine lower and upper thresholds and retain only data that falls within these ranges. The specific thresholds for each language are reported in Table~\ref{tab:range}.

\paragraph{NER-based Filtering} We adopt language-specific NER models for four languages: English, Arabic, Hindi and Turkish. Specifically, we use spaCy model \textit{en\_core\_web\_sm} for English, IndicNER for Hindi \cite{mhaske2022naamapadam}, the CAMeLBERT MSA NER Model for Arabic \cite{inoue-etal-2021-interplay} and the Bert-base-turkish-cased model\footnote{https://huggingface.co/akdeniz27/bert-base-turkish-cased-ner} for Turkish. Named entities identified by these models are subsequently tokenized using the tokenizer. We then exclude samples where named entities constitute more than 60\% of the total token length.

\begin{table}
  \centering
  \begin{tabular}{cc}  
  \hline
    \textbf{Language} & \textbf{Length Threshold} \\
    \hline
    Arabic & [20, 80] \\
    Bengali & [50, 250] \\
    Bulgarian & [20, 140] \\
    Chinese & [10, 150] \\
    Czech & [20, 120] \\
    Dutch & [20, 100] \\
    English & [10, 50] \\
    Finnish & [20, 100] \\
    French & [10, 120] \\
    German & [20, 90] \\
    Hindi & [50, 230] \\
    Hungarian & [20, 120] \\
    Icelandic & [20, 110] \\
    Indonesian & [10, 100] \\
    Italian & [20, 100] \\
    Macedonian & [30, 120] \\
    Polish & [20, 100] \\
    Portuguese & [20, 100] \\
    Romanian & [20, 100] \\
    Russian & [30, 180] \\
    Spanish & [10, 100] \\
    Turkish & [20, 80] \\
    Ukrainian & [20, 150] \\
    \hline
  \end{tabular}
  \caption{The length range we adopt for different languages.}
  \label{tab:range}
\end{table}

\section{Evaluation Details}
\label{sec:eval_detail}
We use the BenchMAX evaluation suite for all the models and language directions. The decoding strategy is greedy decoding for LLM-based decoder-only models and beam search for NLLB-200-1.3B (beam size=5, length penalty=0.6).
For LLaMAX3-8B-Alpaca, both evaluation and training use the prompt described in the original work to maintain consistency. The full prompt template is provided below.



\begin{tcolorbox}[
    colframe=gray!80!black, 
    colback=gray!10!white, 
    coltitle=white, 
    fonttitle=\bfseries, 
    title=Template for LLaMAX\label{long_open_q}, 
    boxrule=0.5mm, 
]

        User: Below is an instruction that describes a task, paired with an input that provides further context.\\
        Write a response that appropriately completes the request.\\
        \#\#\# Instruction:\\
        Translate the following sentences from \{src\_lang\} to \{tgt\_lang\}.\\
        \#\#\# Input:\\
        \{src\_text\}\\
        \#\#\# Response: \\
        Assistant:
\label{prompt:llamax}
\end{tcolorbox}

\section{LLM-as-a-Judge Prompt}
\label{appendix:llm-as-judge}
In Table~\ref{tab:main_tab}, we use LLM-as-a-Judge to evaluate the translation quality of different models. We adopt the ESA-like prompt from \cite{lavie-etal-2025-findings} and add a human reference in the prompt to further improve the evaluation accuracy of LLM-as-a-Judge.

\begin{tcolorbox}[
    colframe=gray!80!black, 
    colback=gray!10!white, 
    coltitle=white, 
    fonttitle=\bfseries, 
    title=LLM-as-a-Judge Prompt\label{long_open_q}, 
    boxrule=0.5mm, 
]

        Score the following translation from \{source\_lang\} to \{target\_lang\} with respect to the human reference on a scale from 0 to 100, where a score of 0 means a broken or poor translation; 33 indicates a flawed translation with significant issues; 66 indicates a good translation with only minor issues in grammar, fluency, or consistency; and 100 represents a perfect translation in both meaning and grammar. Answer with only a whole number representing the score, and nothing else. \\
        \{source\_lang\} source text: \\
        \{source\_seg\} \\
        \{target\_lang\} reference:\\
        \{reference\_seg\} \\
        \{target\_lang\} translation:\\
        \{target\_seg\}
\label{prompt:llamax}
\end{tcolorbox}

\section{Additional Results on FLORES-101}
\label{app:borda}
We report the average rank of each model in Table~\ref{tab:borda}.

\begingroup
\renewcommand{\arraystretch}{1} 
\begin{table*}[!ht]
\centering
\footnotesize
\resizebox{\linewidth}{!}{
\begin{tabular}{l|cccccccccccccc}

\toprule
& \multicolumn{1}{c}{\textbf{x $\rightarrow$ en}} & \multicolumn{1}{c}{\textbf{en $\rightarrow$ x}} & \multicolumn{1}{c}{\textbf{x $\rightarrow$ ar}} & \multicolumn{1}{c}{\textbf{ar $\rightarrow$ x}} & \multicolumn{1}{c}{\textbf{x $\rightarrow$ tr}} & \multicolumn{1}{c}{\textbf{tr $\rightarrow$ x}} & \multicolumn{1}{c}{\textbf{x $\rightarrow$ hi}} & \multicolumn{1}{c}{\textbf{hi $\rightarrow$ x}} & \multicolumn{1}{c}{\textbf{x $\rightarrow$ ru}} & \multicolumn{1}{c}{\textbf{ru $\rightarrow$ x}} & \multicolumn{1}{c}{\textbf{x $\rightarrow$ zh}} & \multicolumn{1}{c}{\textbf{zh $\rightarrow$ x}} & \multicolumn{1}{c}{\textbf{x $\rightarrow$ sw}} & \multicolumn{1}{c}{\textbf{sw $\rightarrow$ x}} \\
& \textbf{spBLEU} & \textbf{spBLEU} & \textbf{spBLEU} & \textbf{spBLEU} & \textbf{spBLEU} & \textbf{spBLEU} & \textbf{spBLEU} & \textbf{spBLEU} & \textbf{spBLEU} & \textbf{spBLEU} & \textbf{spBLEU} & \textbf{spBLEU} & \textbf{spBLEU} & \textbf{spBLEU} \\

\midrule
Hunyuan-MT-7B  
& 9.317 & 6.505 & 4.812 & 6.604 & 4.475 & 6.208 & 7.762 & 6.921 & 8.683 & 6.604 & 8.446 & 5.782 & 5.762 & 6.950 \\

Tower-Plus-9B 
& 3.238 & 6.762 & 9.238 & 6.604 & 7.861 & 6.495 & 2.386 & 6.208 & 2.901 & 6.584 & 2.515 & 6.594 & 7.624 & 6.366 \\

Aya-Expanse-8B 
& 7.020 & 6.426 & 6.426 & 6.089 & 7.158 & 6.158 & 6.436 & 6.307 & 6.842 & 6.228 & 7.683 & 6.218 & 8.782 & 8.337 \\

\midrule
Qwen3-8B
& 4.941 & 7.020 & 6.089 & 7.386 & 7.119 & 7.515 & 8.713 & 7.455 & 7.079 & 7.396 & 3.475 & 6.921 & 9.851 & 8.079 \\

\textbf{+\method}
& 6.050 & 6.465 & 6.822 & 6.376 & 6.228 & 6.386 & 8.901 & 6.752 & 7.257 & 6.683 & 4.069 & 6.168 & 7.455 & 6.624 \\

\midrule
Translategemma-4b-it 
& 7.515 & 6.000 & 2.980 & 6.248 & 3.950 & 6.178 & 3.525 & 6.059 & 5.594 & 6.059 & 5.525 & 6.356 & 4.891 & 6.099 \\

\textbf{+\method} 
& 6.515 & 4.287 & 2.792 & 4.554 & 3.475 & 4.406 & 3.851 & 4.257 & 4.238 & 4.406 & 4.752 & 5.069 & 3.465 & 3.970 \\

\midrule
LLaMAX3-8B-Alpaca 
& 4.129 & 5.168 & 5.802 & 5.257 & 8.594 & 5.485 & 5.881 & 5.040 & 5.525 & 4.901 & 7.614 & 5.416 & 3.861 & 4.139 \\

\textbf{+W-SFT}
& 2.762 & 3.693 & 4.485 & 3.307 & 3.960 & 3.871 & 3.455 & 3.653 & 3.733 & 3.317 & 6.178 & 3.713 & 2.099 & 2.861 \\

\textbf{+\method}  
& 3.515 & 2.673 & 5.554 & 2.574 & 2.178 & 2.297 & 4.089 & 2.347 & 3.149 & 2.822 & 4.743 & 2.762 & 1.208 & 1.574 \\
\bottomrule

& \multicolumn{1}{c}{\textbf{x $\rightarrow$ en}} & \multicolumn{1}{c}{\textbf{en $\rightarrow$ x}} & \multicolumn{1}{c}{\textbf{x $\rightarrow$ ar}} & \multicolumn{1}{c}{\textbf{ar $\rightarrow$ x}} & \multicolumn{1}{c}{\textbf{x $\rightarrow$ tr}} & \multicolumn{1}{c}{\textbf{tr $\rightarrow$ x}} & \multicolumn{1}{c}{\textbf{x $\rightarrow$ hi}} & \multicolumn{1}{c}{\textbf{hi $\rightarrow$ x}} & \multicolumn{1}{c}{\textbf{x $\rightarrow$ ru}} & \multicolumn{1}{c}{\textbf{ru $\rightarrow$ x}} & \multicolumn{1}{c}{\textbf{x $\rightarrow$ zh}} & \multicolumn{1}{c}{\textbf{zh $\rightarrow$ x}} & \multicolumn{1}{c}{\textbf{x $\rightarrow$ sw}} & \multicolumn{1}{c}{\textbf{sw $\rightarrow$ x}} \\
& \textbf{xCOMET*} & \textbf{xCOMET*} & \textbf{xCOMET*} & \textbf{xCOMET*} & \textbf{xCOMET*} & \textbf{xCOMET*} & \textbf{xCOMET*} & \textbf{xCOMET*} & \textbf{xCOMET*} & \textbf{xCOMET*} & \textbf{xCOMET*} & \textbf{xCOMET*} & \textbf{xCOMET*} & \textbf{xCOMET*} \\

\midrule
Hunyuan-MT-7B  
& 6.938 & 5.700 & 5.450 & 5.750 & 3.825 & 5.875 & 4.188 & 6.125 & 5.350 & 5.950 & 4.325 & 5.825 & 6.350 & 7.912 \\

Tower-Plus-9B 
& 5.275 & 6.700 & 9.650 & 6.987 & 9.150 & 6.875 & 3.812 & 6.600 & 4.388 & 6.662 & 5.362 & 6.725 & 7.950 & 7.225 \\

Aya-Expanse-8B 
& 7.675 & 7.050 & 6.487 & 6.550 & 8.200 & 6.600 & 7.362 & 6.725 & 7.700 & 7.037 & 8.488 & 6.900 & 8.625 & 8.838 \\

\midrule
Qwen3-8B
& 6.900 & 7.513 & 7.862 & 7.487 & 8.037 & 7.662 & 9.338 & 7.625 & 8.700 & 7.487 & 5.250 & 7.412 & 9.838 & 7.987 \\

\textbf{+\method}
& 5.225 & 5.912 & 5.850 & 6.025 & 5.987 & 6.125 & 8.137 & 6.175 & 6.800 & 6.100 & 3.513 & 5.838 & 7.225 & 6.812 \\

\midrule
Translategemma-4b-it 
& 5.275 & 4.487 & 5.375 & 4.963 & 3.587 & 5.125 & 2.688 & 5.125 & 3.513 & 4.750 & 4.062 & 4.912 & 5.013 & 4.537 \\

\textbf{+\method} 
& 4.375 & 3.263 & 4.588 & 3.625 & 2.275 & 3.663 & 3.413 & 3.337 & 3.163 & 3.587 & 4.350 & 3.650 & 3.837 & 3.163 \\

\midrule
LLaMAX3-8B-Alpaca 
& 6.263 & 6.100 & 5.650 & 5.912 & 6.575 & 5.825 & 6.625 & 5.862 & 6.562 & 5.812 & 7.763 & 6.013 & 3.138 & 4.112 \\

\textbf{+W-SFT} 
& 5.000 & 4.812 & 3.087 & 4.562 & 4.775 & 4.487 & 5.388 & 4.425 & 5.300 & 4.537 & 6.487 & 4.600 & 2.025 & 2.850 \\

\textbf{+\method}  
& 2.075 & 3.288 & 1.000 & 2.950 & 2.587 & 2.625 & 4.050 & 2.850 & 3.525 & 2.950 & 5.400 & 2.975 & 1.000 & 1.387 \\
\bottomrule

& \multicolumn{1}{c}{\textbf{x $\rightarrow$ en}} & \multicolumn{1}{c}{\textbf{en $\rightarrow$ x}} & \multicolumn{1}{c}{\textbf{x $\rightarrow$ ar}} & \multicolumn{1}{c}{\textbf{ar $\rightarrow$ x}} & \multicolumn{1}{c}{\textbf{x $\rightarrow$ tr}} & \multicolumn{1}{c}{\textbf{tr $\rightarrow$ x}} & \multicolumn{1}{c}{\textbf{x $\rightarrow$ hi}} & \multicolumn{1}{c}{\textbf{hi $\rightarrow$ x}} & \multicolumn{1}{c}{\textbf{x $\rightarrow$ ru}} & \multicolumn{1}{c}{\textbf{ru $\rightarrow$ x}} & \multicolumn{1}{c}{\textbf{x $\rightarrow$ zh}} & \multicolumn{1}{c}{\textbf{zh $\rightarrow$ x}} & \multicolumn{1}{c}{\textbf{x $\rightarrow$ sw}} & \multicolumn{1}{c}{\textbf{sw $\rightarrow$ x}} \\
& \textbf{MetricX*} & \textbf{MetricX*} & \textbf{MetricX*} & \textbf{MetricX*} & \textbf{MetricX*} & \textbf{MetricX*} & \textbf{MetricX*} & \textbf{MetricX*} & \textbf{MetricX*} & \textbf{MetricX*} & \textbf{MetricX*} & \textbf{MetricX*} & \textbf{MetricX*} & \textbf{MetricX*} \\

\midrule
Hunyuan-MT-7B  
& 6.679 & 6.429 & 6.548 & 6.393 & 4.357 & 6.369 & 5.214 & 6.810 & 5.333 & 6.452 & 4.452 & 6.000 & 6.107 & 7.905 \\

Tower-Plus-9B 
& 5.298 & 6.905 & 9.095 & 7.155 & 8.750 & 7.036 & 4.560 & 6.857 & 5.000 & 6.869 & 6.095 & 6.512 & 7.036 & 6.833 \\

Aya-Expanse-8B 
& 8.071 & 7.143 & 8.131 & 7.131 & 8.500 & 7.000 & 8.071 & 7.095 & 8.131 & 7.262 & 8.833 & 6.869 & 9.940 & 8.798 \\

\midrule
Qwen3-8B 
& 6.881 & 7.452 & 7.690 & 7.583 & 7.750 & 7.595 & 8.905 & 7.607 & 8.226 & 7.476 & 5.738 & 7.024 & 9.036 & 7.952 \\

\textbf{+\method} 
& 5.345 & 5.798 & 6.262 & 5.869 & 5.714 & 6.060 & 7.833 & 5.917 & 6.500 & 5.952 & 4.238 & 5.226 & 7.714 & 6.690 \\

\midrule
Translategemma-4b-it 
& 5.048 & 4.131 & 2.369 & 4.381 & 3.952 & 4.345 & 3.119 & 4.512 & 3.738 & 4.738 & 4.226 & 6.071 & 4.024 & 3.893 \\

\textbf{+\method} 
& 4.702 & 2.726 & 2.429 & 2.548 & 2.893 & 2.893 & 2.893 & 2.417 & 3.167 & 2.750 & 3.238 & 5.298 & 2.607 & 2.750 \\

\midrule
LLaMAX3-8B-Alpaca 
& 5.679 & 5.976 & 5.976 & 5.798 & 6.333 & 5.869 & 6.357 & 5.821 & 6.345 & 5.679 & 7.298 & 5.274 & 4.143 & 4.298 \\

\textbf{+W-SFT} 
& 4.560 & 4.643 & 4.190 & 4.440 & 4.357 & 4.464 & 4.750 & 4.452 & 5.131 & 4.310 & 6.179 & 3.714 & 2.905 & 3.179 \\

\textbf{+\method}  
& 2.738 & 3.190 & 2.310 & 2.964 & 2.393 & 2.762 & 3.286 & 2.940 & 3.417 & 2.905 & 4.702 & 2.321 & 1.488 & 1.857 \\
\bottomrule

\end{tabular}%
}

\caption{Average rank of strong multilingual LLMs on the FLORES-101 test set, with results for 7 representative languages shown in the table.}
\label{tab:borda}
\end{table*}
\endgroup

\section{More cases of Holes in Machine Translation Metrics}

\label{app:hole_examples}
More failure cases of MetricX are shown in Figure~\ref{app:case1}, Figure~\ref{app:case2}, Figure~\ref{app:case3} and Figure~\ref{app:case4}.
Together, these examples show that the holes in QE are versatile and lead to reward hacking during reinforcement training.


\section{Qualitative Examples of \method}
\label{app:qualitative_examples}
We add qualitative translation examples illustrating how \method improves LLMs' translation quality relative to the baselines in Figure~\ref{app:case5}, Figure~\ref{app:case6}, Figure~\ref{app:case7} and Figure~\ref{app:case8}, with xCOMET scores provided for reference.

\section{Human Evaluation}
\label{app:human_eval}
\begin{figure*}[!htb]
    \centering
    \includegraphics[width=1\textwidth]{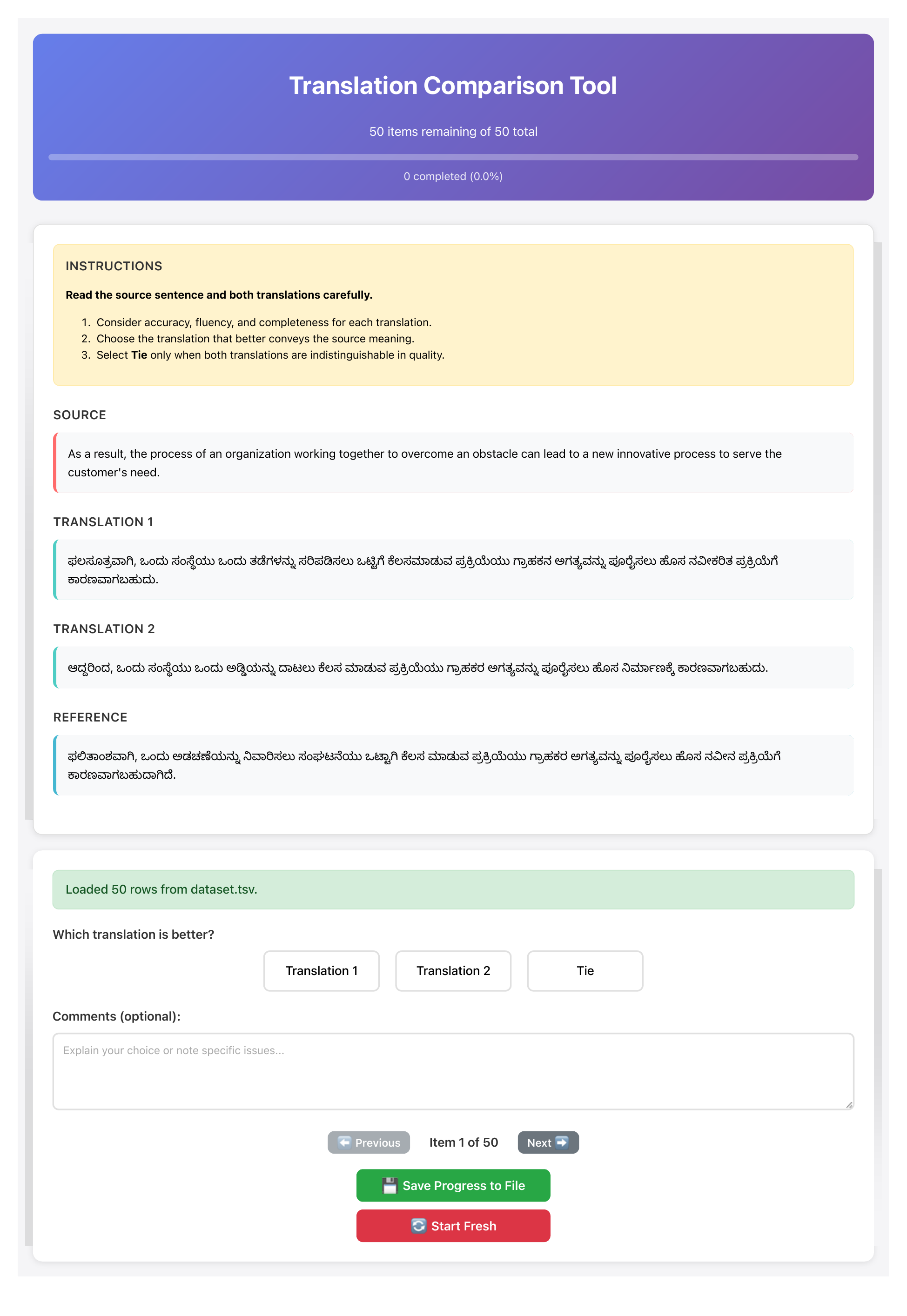}
    \vspace{-1em}
    \caption{Screenshot of human evaluation web tool.}
    \label{fig:annotate_html}
\end{figure*}

We hired native speakers in the university lab to serve as human annotators and compensated them at the U.S. minimum wage. We provide the screenshot of our annotation page in Figure \ref{fig:annotate_html}.

\section{Training Languages}
\label{app:lang_info}
 In total, our training dataset covers 22 source languages (Arabic, Bengali, Bulgarian, Croatian, German, English, Finnish, French, Hindi, Hungarian, Indonesian, Italian, Icelandic, Macedonian, Dutch, Polish, Portuguese, Romanian, Russian, Spanish, Turkish, Ukrainian, Simple Chinese.) and 1,016 language directions. We remove target languages that are either not supported by MetricX or not segmented by spaces   (except for Simple Chinese and Traditional Chinese, for which we use HanLP\footnote{https://github.com/hankcs/HanLP} to tokenize the sentence). For each direction, we sample 250 instances and train all language directions concurrently. This makes our training dataset consist of 254,000 monolingual sentences in total.

\section{Used Scientific Artifacts}
Below are the scientific artifacts we've used in our paper. For the sake of ethics, we ensure all usages comply with their license.
\begin{itemize}
    \item \textit{OpenRLHF (Apache-2.0 license)}, an open-source RLHF framework that integrates high performance with simple usage, aiming to streamline the training process and enhance the accessibility of RLHF methods.
    \item \textit{spaCy (MIT license)}, a library for advanced Natural Language Processing in Python and Cython, build on the very latest research, and was designed to be used in real products.
    \item \textit{vLLM (Apache-2.0 license)}, a fast and easy-to-use library optimized specifically for LLM inference and serving.
    \item \textit{Transformers (Apache-2.0 license)}, a model-definition framework focusing on machine learning models for both inference and training.
\end{itemize}

\begin{figure*}[t]
\vspace{-0.2cm}
\footnotesize
\centering
\begin{tabular}{ |p{\dimexpr\linewidth-4\tabcolsep-1.5pt}| }
\toprule

\multicolumn{1}{|p{\dimexpr\linewidth-4\tabcolsep-1.5pt}|}{
{\textbf{Source (English):} Dr. Tony Moll discovered the Extremely Drug Resistant Tuberculosis (XDR-TB) in the South African region KwaZulu-Natal.}
} \\
\midrule
{
\begin{CJK}{UTF8}{gbsn}
\textbf{Reference (Spanish):} La tuberculosis extremadamente resistente (XDR-TB) fue detectada por primera vez por el Dr. Tony Moll en \'area sudafricana de Zululandia.
\end{CJK}
}
\\
\midrule
{
\begin{CJK}{UTF8}{gbsn}
\textbf{Over-translation (Spanish):}
El Dr. Tony Moll descubri\'o en la regi\'on de KwaZulu-Natal, en Sud\'africa, un tipo de tuberculosis muy difícil de tratar: la tuberculosis extremadamente resistente a los medicamentos (XDR-TB). {\color{red} Esta bacteria es resistente a la mayoría de los tratamientos antibi\'oticos convencionales y requiere tratamiento con medicamentos antituberculosos específicos.} {\color{blue}MetricX: -3.11}
\end{CJK}
} \\
\midrule
{\textbf{Wrong Language (Portuguese):} 
 {\color{red}O Dr. Tony Moll descobriu a Tuberculose Extremamente Resistente a Drogas (XDR-TB) na regi\~ao de KwaZulu-Natal, na \'africa do Sul.} {\color{blue}MetricX: -1.85}}
\\
\bottomrule
\end{tabular}
\caption{
 A case study from \textsc{Flores-101} dataset. The intended language direction is from English to Spanish. {\color{blue}blue} text denotes the MetricX score in source-based mode, and the {\color{red}red} text highlights the errors in the translation.
}
\vspace{-0.3cm}
\label{app:case1}
\end{figure*}

\begin{figure*}[t]
\vspace{-0.2cm}
\footnotesize
\centering
\begin{tabular}{ |p{\dimexpr\linewidth-4\tabcolsep-1.5pt}| }
\toprule

\multicolumn{1}{|p{\dimexpr\linewidth-4\tabcolsep-1.5pt}|}{
{\textbf{Source (English):} Because the dinosaur feathers do not have a well-developed shaft, called a rachis, but do have other features of feathers - barbs and barbules - the researchers inferred the rachis was likely a later evolutionary development that these other features.}
} \\
\midrule
{
\begin{CJK}{UTF8}{gbsn}
\textbf{Reference (Polish):} W związku z tym, że pi\'ora dinozaura nie mają dobrze rozwiniętego trzonu, zwanego stosiną, lecz mają inne cechy pi\'or – promienie i promyki – naukowcy wywnioskowali, że stosina prawdopodobnie była p\'oźniejszym etapem ewolucyjnym niż te pozostałe cechy.
\end{CJK}
}
\\
\midrule
{
\begin{CJK}{UTF8}{gbsn}
\textbf{Over-translation (Polish):}
Ponieważ pi\'ora dinozaur\'ow nie posiadały typowej osi (centralnego trzonu przebiegającego przez całą strukturę pi\'ora), ale zachowały podstawowe cechy pi\'or, takie jak promienie i promyczki, naukowcy wysnuli wniosek, że struktura osi mogła ewoluować p\'oźniej, podczas gdy {\color{red}inne cechy, takie jak promienie i promyczki, mogły istnieć już wcześniej.} {\color{blue}MetricX: -3.17}
\end{CJK}
} \\
\midrule
{\textbf{Wrong Language (Slovak):} 
 {\color{red}Preto\v{z}e dinosaurov\'e perie nem\'a dobre vyvinut\'u osku, naz\'yvan\'u rachis, ale m\'a in\'e \v{c}rty peria - h\'a\v{c}iky a h\'a\v{c}ikovit\'e vl\'akna, vedci us\'udili, \v{z}e rachis pravdepodobne vznikol a\v{z} nesk\^or v evol\'ucii ne\v{z} tieto ostatn\'e \v{c}rty.} {\color{blue}MetricX: -3.52}} 
\\
\bottomrule
\end{tabular}
\caption{
 A case study from \textsc{Flores-101} dataset. The intended language direction is English to Polish. {\color{blue}Blue} text denotes the MetricX score in source-based mode, and the {\color{red}red} text highlights the errors in the translation.
}
\vspace{-0.3cm}
\label{app:case2}
\end{figure*}

\begin{figure*}[t]
\vspace{-0.2cm}
\footnotesize
\centering
\begin{tabular}{ |p{\dimexpr\linewidth-4\tabcolsep-1.5pt}| }
\toprule

\multicolumn{1}{|p{\dimexpr\linewidth-4\tabcolsep-1.5pt}|}{
{\textbf{Source (English):} Previously, Ring's CEO, Jamie Siminoff, remarked the company started when his doorbell wasn't audible from his shop in his garage. Translate it from English to Chinese.}
} \\
\midrule
{
\begin{CJK}{UTF8}{gbsn}
\textbf{Reference (Chinese):} 此前，铃声 (Ring) 公司 CEO 杰米·西米诺夫曾表示，公司成立之初，他在位于车库的工作室里是听不到门铃响声的。
\end{CJK}
}
\\
\midrule
{
\begin{CJK}{UTF8}{gbsn}
\textbf{Over-translation (Chinese):}
以前，Ring公司的CEOJamie Siminoff提到，公司最初成立时，{\color{red}其主要业务是通过门铃来吸引顾客。}他的店铺位于车库内，而门铃则安装在车库的入口处。 {\color{blue}MetricX: -3.70}
\end{CJK}
} \\
\midrule
{\textbf{Wrong Language (Japanese):} 
\begin{CJK}{UTF8}{min}
 {\color{red}以前、RingのCEOであるジェイミー・シミノフ氏は、自宅のガレージにある作業場からドアベルの音が聞こえなかったことがきっかけで同社を創業したと述べました。} {\color{blue}MetricX: -3.59}
\end{CJK}} 
\\
\midrule
{
\begin{CJK}{UTF8}{gbsn}
\textbf{Non-translation (English):}
{\color{red}Earlier, Jamie Siminoff, CEO of Ring, shared the story behind the company’s founding; it all began when he noticed that the doorbell at his house wasn’t loud enough for him to hear it clearly from his workshop, which was located in his garage.} {\color{blue}MetricX: -0.66}
\end{CJK}
} \\
\bottomrule
\end{tabular}
\caption{
 A case study from \textsc{Flores-101} dataset. The intended language direction is from English to Chinese. {\color{blue}Blue} text denotes the MetricX score in source-based mode, and the {\color{red}red} text highlights the errors in the translation.
}
\vspace{-0.3cm}
\label{app:case3}
\end{figure*}

\begin{figure*}[t]
\vspace{-0.2cm}
\footnotesize
\centering
\begin{tabular}{ |p{\dimexpr\linewidth-4\tabcolsep-1.5pt}| }
\toprule

\multicolumn{1}{|p{\dimexpr\linewidth-4\tabcolsep-1.5pt}|}{
{\textbf{Source (French):} Les travailleurs doivent souvent obtenir l'approbation de leurs sup\'erieurs pour toute d\'ecision qu'ils prennent et sont cens\'es ob\'eir aux instructions de ces derniers sans les remettre en question.}
} \\
\midrule
{
\begin{CJK}{UTF8}{gbsn}
\textbf{Reference (German):} Oftmals müssen Arbeitnehmer für alle Entscheidungen, die sie treffen, die Zustimmung ihrer Vorgesetzten einholen und es wird von ihnen erwartet, dass sie den Anweisungen ihrer Chefs bedingungslos Folge leisten.
\end{CJK}
}\\
\midrule
{
\begin{CJK}{UTF8}{gbsn}
\textbf{Over-translation (German):}
Die Mitarbeiter müssen vor jeder Entscheidung die Zustimmung ihrer Vorgesetzten einholen und sich strikt an deren Anweisungen halten, {\color{red}ohne diese willkürlich zu ändern oder deren Ausführung zu verweigern.} {\color{blue}MetricX: -0.29}
\end{CJK}
} \\
\midrule
{\textbf{Wrong Language (Italian):} 
\begin{CJK}{UTF8}{min}
 {\color{red} I lavoratori devono spesso ottenere l’approvazione dei loro superiori per qualsiasi decisione prendano e sono tenuti a seguire le loro istruzioni senza metterle in discussione.} {\color{blue}MetricX: -0.35}
\end{CJK}} 
\\
\midrule
{
\begin{CJK}{UTF8}{gbsn}
\textbf{Non-translation (English):}
{\color{red}Les employ\'es sont g\'en\'eralement tenus de solliciter l’accord de leurs responsables pour chaque d\'ecision et doivent suivre leurs directives sans les contester.}{\color{blue}MetricX: -0.37}
\end{CJK}
} \\
\bottomrule
\end{tabular}
\caption{
 A case study from \textsc{Flores-101} dataset. The intended language direction is from French to German. {\color{blue}Blue} text denotes the MetricX score in source-based mode, and the {\color{red}red} text highlights the errors in the translation.
}
\vspace{-0.3cm}
\label{app:case4}
\end{figure*}

\begin{figure*}[t]
\vspace{-0.2cm}
\footnotesize
\centering
\begin{tabular}{ |p{\dimexpr\linewidth-4\tabcolsep-1.5pt}| }
\toprule

\multicolumn{1}{|p{\dimexpr\linewidth-4\tabcolsep-1.5pt}|}{
{\textbf{Source (English):} In late 2017, Siminoff appeared on shopping television channel QVC.}
} \\
\midrule
{
\begin{CJK}{UTF8}{gbsn}
\textbf{Reference (Xhosa):} Ngasekupheleni kuka-2017, uSiminoff kwitshaneli yokuthenga kwiTV enguQVC.
\end{CJK}
}
\\
\midrule
{
\begin{CJK}{UTF8}{gbsn}
\textbf{LLaMAX3-8B-Alpaca (Xhosa):}
Kwiyeza zokuthenga zesikhathi zesibini ze-2017, uSiminoff wabonakala kwiQVC, isiteshi se-TV seziyobisi. {\color{blue}xCOMET: 49.52}
\end{CJK}
} \\
\midrule
{\textbf{LLaMAX3-8B-Alpaca+\method (Xhosa):} 
 Ngasekupheleni kuka-2017, uSiminoff waboniswa kwinkqubo yentengiso yeQVC, isiteshi seTV.    {\color{blue}xCOMET: 82.14}} 
\\
\bottomrule
\end{tabular}
\caption{
 Case study of the improvement brought by \method. The intended language direction is from English to Xhosa. {\color{blue}Blue} text denotes the xCOMET score in reference-based mode.
}
\vspace{-0.3cm}
\label{app:case5}
\end{figure*}

\begin{figure*}[t]
\vspace{-0.2cm}
\footnotesize
\centering
\begin{tabular}{ |p{\dimexpr\linewidth-4\tabcolsep-1.5pt}| }
\toprule

\multicolumn{1}{|p{\dimexpr\linewidth-4\tabcolsep-1.5pt}|}{
{\textbf{Source (English):} One antibody cocktail, ZMapp, initially showed promise in the field, but formal studies indicated it had less benefit than sought in preventing death.}
} \\
\midrule
{
\begin{CJK}{UTF8}{gbsn}
\textbf{Reference (Chinese):} 一种名为 ZMapp 的“抗体鸡尾酒”（多抗体联合）最初有望在该领域发挥作用，但正式研究表明，它在预防死亡方面的效果有些不尽人意。
\end{CJK}
}
\\
\midrule
{
\begin{CJK}{UTF8}{gbsn}
\textbf{LLaMAX3-8B-Alpaca (Chinese):}
一种抗体混合物，ZMapp，最初显示出在现场的希望，但正式研究表明它比预期的死亡率降低。 {\color{blue}xCOMET: 58.56}
\end{CJK}
} \\
\midrule
\begin{CJK}{UTF8}{gbsn}
{\textbf{LLaMAX3-8B-Alpaca+\method (Chinese):} 
 一种抗体混合物，ZMapp，最初显示出希望，但正式研究表明它无法达到预期的防止死亡效果。    {\color{blue}xCOMET: 75.04}}
\end{CJK}
\\
\bottomrule
\end{tabular}
\caption{
 Case study of the improvement brought by \method. The intended language direction is English to Chinese. {\color{blue}Blue} text denotes the xCOMET score in reference-based mode.
}
\vspace{-0.3cm}
\label{app:case6}
\end{figure*}

\begin{figure*}[t]
\vspace{-0.2cm}
\footnotesize
\centering
\begin{tabular}{ |p{\dimexpr\linewidth-4\tabcolsep-1.5pt}| }
\toprule

\multicolumn{1}{|p{\dimexpr\linewidth-4\tabcolsep-1.5pt}|}{
\begin{CJK}{UTF8}{gbsn}
{\textbf{Source (Chinese):} 在 PALM 实验中，ZMapp 用作参照。也就是科学家将其作为基线，把其他三种治疗方法与之作比较。}
\end{CJK}
} \\
\midrule
{
\begin{CJK}{UTF8}{gbsn}
\textbf{Reference (Swahili):} Katika jaribio la PALM, Zmapp ilitumika kama kidhibiti, kumaanisha wanasayansi waliitumia kama msingi na kulinganisha mat
ibabu mengine matatu nayo.
\end{CJK}
}
\\
\midrule
{
\begin{CJK}{UTF8}{gbsn}
\textbf{LLaMAX3-8B-Alpaca (Swahili):}
Katika majaribio ya Palm, ZMapp hutumiwa kama kumbukumbu. Hivyo, wanasayansi wamefanya kulinganisha kati ya matibabu mengine matatu na yake. Kwa kufanya hivyo, wanasayansi wanaweza kuchunguza jinsi matibabu mengine yanavyofanya kazi kwa kulinganisha na ZMapp. {\color{blue}xCOMET: 65.48}
\end{CJK}
} \\
\midrule
\begin{CJK}{UTF8}{gbsn}
{\textbf{LLaMAX3-8B-Alpaca+\method (Swahili):} 
 Katika majaribio ya PALM, ZMapp ilitumiwa kama kipimo cha kulinganisha; yaani, wataalamu walitumia kama kiwango na kulinganisha njia tatu nyingine za matibabu na hiyo.    {\color{blue}xCOMET: 93.80}}
\end{CJK}
\\
\bottomrule
\end{tabular}
\caption{
 Case study of the improvement brought by \method. The intended language direction is from Chinese to Swahili. {\color{blue}Blue} text denotes the xCOMET score in reference-based mode.
}
\vspace{-0.3cm}
\label{app:case7}
\end{figure*}

\begin{figure*}[t]
\vspace{-0.2cm}
\footnotesize
\centering
\begin{tabular}{ |p{\dimexpr\linewidth-4\tabcolsep-1.5pt}| }
\toprule

\multicolumn{1}{|p{\dimexpr\linewidth-4\tabcolsep-1.5pt}|}{
\begin{CJK}{UTF8}{gbsn}
{\textbf{Source (Chinese):} 西米诺夫说，2013 年他在《创智赢家》节目中露面后，公司的销售额大增，当时节目组拒绝向这家初创公司投资。}
\end{CJK}
} \\
\midrule
{
\begin{CJK}{UTF8}{gbsn}
\textbf{Reference (Swahili):} M. Siminoff a d\'eclar\'e que les ventes ont augment\'e apr\`es son apparition en 2013 dans un \'episode de Shark Tank où le panel de la s\'erie a refus\'e de financer la start-up.
\end{CJK}
}
\\
\midrule
{
\begin{CJK}{UTF8}{gbsn}
\textbf{LLaMAX3-8B-Alpaca (French):}
Westminster a d\'eclar\'e que depuis son apparition dans l'\'emission "Entrepreneurs de g\'enie" en 2013, les ventes de sa soci\'et\'e ont augment\'e consid\'erablement. Cependant, le programme n'a pas investi dans l'entreprise. {\color{blue}xCOMET: 15.52}
\end{CJK}
} \\
\midrule
\begin{CJK}{UTF8}{gbsn}
{\textbf{LLaMAX3-8B-Alpaca+\method (French):} 
 Siminov a d\'eclar\'e que, apr\`es son apparition dans le programme «Entrepreneurs cr\'eatifs» en 2013, les ventes de sa soci\'et\'e ont consid\'erablement augment\'e, mais le programme n'a pas investi dans cette entreprise d\'ebutante.    {\color{blue}xCOMET: 45.24}}
\end{CJK}
\\
\bottomrule
\end{tabular}
\caption{
 Case study of the improvement brought by \method. The intended language direction is from Chinese to French. {\color{blue}Blue} text denotes the xCOMET score in reference-based mode.
}
\vspace{-0.3cm}
\label{app:case8}
\end{figure*}

\begin{table*}
  \centering
  \begin{tabular}{c|ccccccc}  
  \toprule
    \textbf{Models} & \textbf{en $\rightarrow$ x} & \textbf{sw $\rightarrow$ x} & \textbf{tr $\rightarrow$ x} & \textbf{hi $\rightarrow$ x} & \textbf{zh $\rightarrow$ x} & \textbf{ar $\rightarrow$ x} & \textbf{ru $\rightarrow$ x}\\
    \hline
    LLaMAX3-8B-Alpaca & 0.9027 & 0.8931 & 0.8975 & 0.8931 & 0.8969 & 0.9018 & 0.8996 \\
    Aya-Expanse-8B & 0.7481 & 0.7440 & 0.7604 & 0.7537 & 0.7387 & 0.7563 & 0.7440 \\
    Tower-Plus-9B & 0.6199 & 0.7133& 0.6504& 0.5908& 0.6537& 0.7063& 0.6432 \\
    \method & \textbf{0.9270}& \textbf{0.9224}& \textbf{0.9252}& \textbf{0.9192}& \textbf{0.9258}& \textbf{0.9275}&\textbf{0.9273} \\
    \toprule
    \textbf{Models} & \textbf{x $\rightarrow$ en} & \textbf{x $\rightarrow$ sw} & \textbf{x $\rightarrow$ tr} & \textbf{x $\rightarrow$ hi} & \textbf{x $\rightarrow$ zh} & \textbf{x $\rightarrow$ ar} & \textbf{x $\rightarrow$ ru}\\
    \hline
    LLaMAX3-8B-Alpaca & 0.9970& 0.8329& 0.9719& 0.9513& 0.9780& 0.8329& 0.9890\\
    Aya-Expanse-8B & 0.9919& 0.8493& 0.9672& 0.9770& 0.9472& 0.8792& 0.9865 \\
    Tower-Plus-9B &0.9923& 0.7518& 0.9598& 0.9662& 0.9675& 0.8072& 0.9666 \\
    \method & \textbf{0.9996}& \textbf{0.9619}& \textbf{0.9986}& \textbf{0.9913}& \textbf{0.9887}& \textbf{0.9652}& \textbf{0.9986}\\
    \bottomrule
  \end{tabular}
  \caption{Complete results for LCR}
  \label{tab:full_lcr}
\end{table*}

\end{document}